\documentclass[preprint,12pt]{elsarticle}

\usepackage[margin=1in]{geometry} 
\usepackage{natbib}
\usepackage{amsmath}
\usepackage{amsfonts}
\usepackage{amssymb}
\usepackage{graphicx}
\usepackage{graphicx}
\usepackage{bm}
\usepackage{subcaption}
\usepackage{multirow}
\graphicspath{{./}}
\usepackage{soul}
\usepackage{color}
\usepackage[normalem]{ulem}
\usepackage[colorlinks=true,linkcolor=blue,citecolor=blue
]{hyperref}
\usepackage{algorithm}
\usepackage{algpseudocode}
\usepackage{booktabs}
\usepackage{array} 
\newcolumntype{C}[1]{>{\centering\arraybackslash}m{#1}}
\newcolumntype{P}[1]{>{\centering\arraybackslash}p{#1}}

\begin{document}
\begin{frontmatter}

    \title{Bayesian neural networks for predicting uncertainty in full-field material response}
    \author{George D. Pasparakis\corref{cor1}}     
    \cortext[cor1]{Corresponding author.\\E-mail address: \href{mailto:gpaspar1@jh.edu}{gpaspar1@jh.edu} (George D. Pasparakis).}
    \author{Lori Graham-Brady}
    \author{Michael D. Shields}
    \address{Department of Civil and Systems Engineering, Johns Hopkins University, 3400 N. Charles Street, Baltimore, MD 21218, USA}

    \begin{abstract}
        {\it Stress and material deformation field predictions are among the most important tasks in computational mechanics. These predictions are typically made by solving the governing equations of continuum mechanics using finite element analysis, which can become computationally prohibitive considering complex microstructures and material behaviors. Machine learning (ML) methods offer potentially cost effective surrogates for these applications. However, existing ML surrogates are either limited to low-dimensional problems and/or do not provide uncertainty estimates in the predictions. This work proposes an ML surrogate framework for stress field prediction and uncertainty quantification for diverse materials microstructures. A modified Bayesian U-net architecture is employed to provide a data-driven image-to-image mapping from initial microstructure to stress field with prediction (epistemic) uncertainty estimates. The Bayesian posterior distributions for the U-net parameters are estimated using three state-of-the-art inference algorithms: the posterior sampling-based Hamiltonian Monte Carlo method and two variational approaches, the Monte-Carlo Dropout method and the Bayes by Backprop algorithm. A systematic comparison of the predictive accuracy and uncertainty estimates for these methods is performed for a fiber reinforced composite material and polycrystalline microstructure application. It is shown that the proposed methods yield predictions of high accuracy compared to the FEA solution, while uncertainty estimates depend on the inference approach. Generally, the Hamiltonian Monte Carlo and Bayes by Backprop methods provide consistent uncertainty estimates. Uncertainty estimates from Monte Carlo Dropout, on the other hand, are more difficult to interpret and depend strongly on the method's design. 
        } 
        \end{abstract}
\end{frontmatter}

\section{Introduction}
In recent years, the fields of material science and computational mechanics have begun to benefit significantly from advances in machine learning (ML) and data-driven methods. Specifically, computational models in tandem with data-driven methodologies provide promising tools for discovering desirable, but previously unattainable, combinations of features ({\it e.g.}, chemical composition, crystal structure, microstructure) \cite{ludwig2019discovery,bock2019review}. For instance, ML and data-mining have been employed at the atomic level for the construction of complex interatomic potentials \cite{hansen2015machine} and for predicting crystal structure prior to material synthesis \cite{fischer2006predicting}. ML methodologies have also been used to establish structure-material property relationships \cite{kearnes2016molecular} while statistical learning methods have been applied to microstructure characterization \cite{wade2023estimating}, reconstruction \cite{bostanabad2016stochastic} and classification \cite{decost2015computer}. 

In computational mechanics applications, ML techniques have been primarily employed in prediction tasks where the output quantities are relatively low-dimensional. Examples include methods to identify model parameters using regression trees  \cite{yang2018deep}, discover constitutive models \cite{linka2023new}, homogenization of complex material behavior \cite{liu2019deep}, and prediction history-dependent plasticity \cite{mozaffar2019deep} using artificial neural networks. However, a number of works have proposed ML methods for high-dimensional problems, usually relating to full-field modeling and simulation \cite{yabansu2017extraction,zhang2016virtual} where the aim is to evaluate the effects of microstructure morphology on stress and strain localization.In this setting, neural networks (NN) have been increasingly used as surrogates to finite element analysis (FEA), which can become computationally costly for high-dimensional and/or nonlinear problems \cite{reddy2015introduction}. For instance, a hybrid NN-FE model was employed in \cite{hambli2011multiscale} to estimate meso-scale effective bone properties and a convolutional neural network (CNN) was proposed in \cite{gupta2023accelerated} to predict micro-scale stress distributions based on heterogenous material macro-structure. A three-dimensional implementation of this architecture was later used to predict the inelastic behavior of two-phase fiber composites \cite{saha2024prediction}. Further, CNNs have been used to predict the stress field in cantilevered beams \cite{nie2020stress}, fiber reinforced polymers \cite{sun2020predicting} and polycrystalline materials~\cite{mianroodi2021teaching,khorrami2023artificial,frankel2020prediction} and conditional generative adversarial neural networks (cGANs) have been used for stress field estimation in hierarchical composites \cite{yang2021deep}.  In bio-mechanics, fully connected NNs have been used for predicting thoracic aortic stress fields~\cite{liang2018deep}. Graph-based ML methods, which do not require structured data, have also been deployed for predicting stress and strain fields on several material systems~ \cite{maurizi2022predicting}. Further, physics informed neural networks (PINNs) have been used as surrogates for stress field prediction \cite{haghighat2020deep}. Despite their ubiquitous application in two-dimensional problems, the proposed ML surrogates typically provide deterministic estimates without accompanying uncertainty estimates.

Ultimately, a lack of a measurable prediction uncertainty limits the reliability of ML methods as predictive tools in computational mechanics and materials science \cite{bhadeshia2009performance}. In this setting, we must distinguish between aleatoric and epistemic uncertainty~\cite{der2009aleatory}. Aleatoric uncertainty stems from inherent randomness of the process under consideration (e.g., randomness in material properties), which is deemed to be irreducible. Epistemic uncertainty, on the other hand, arises as a result of incomplete understanding or lack of information, which can potentially be reduced by collecting additional data. ML surrogate models for mechanics applications are typically learned from limited and potentially noisy data obtained from expensive-to-evaluate numerical simulations or experiments. This results in a high degree of epistemic uncertainty in the associated models. In addition, ML methods are typically highly overparameterized, resulting in trained models that find solutions in local minima -- thereby introducing epistemic uncertainty in model training. These issues can potentially be mitigated by comprehensive UQ strategies that distinguish between aleatoric and epistemic uncertainty.

Several methods have been developed for surrogate modeling under uncertainty. Among the mostly widely used methods are Gaussian process (GP) regression~\cite{williams2006gaussian} and polynomial chaos expansions (PCE) \cite{xiu2003modeling,kontolati2022manifold}. Notwithstanding their success, these methods do not generally scale well to high-dimensional problems \cite{tripathy2018deep} and often require reducing the dimensionality of the input space \cite{lawrence2003gaussian, giovanis2024polynomial, tripathy2016gaussian}, adding an additional degree of approximation. Recently, there has been renewed interest in the Bayesian treatment of NNs \cite{mackay1992bayesian, kingma2013auto, louizos2017multiplicative} which have the potential to combine the generalizability and accuracy of NNs with robust UQ. However, training Bayesian Neural Networks (BNNs) requires solving an extremely large Bayesian inference problem to learn the distributions of thousands of parameters in the NN. Several approaches to solving this problem have been proposed in the literature and broadly include, but are not limited to: posterior sampling methods based on Markov Chain Monte Carlo (MCMC) \cite{neal2011mcmc, betancourt2017conceptual}, variational inference (VI) methods \cite{hernandez2016black,graves2011practical,gal2016uncertainty,kingma2013auto}, approaches using Laplace approximations \cite{denker1990transforming} and deep ensembles \cite{lakshminarayanan2017simple,izmailov2018averaging}. For a detailed review of the topic the reader is referred to recent review papers \cite{psaros2023uncertainty,jospin2022hands, gawlikowski2023survey}.

Although the construction of BNNs is conceptually straightforward -- parameters of the model are treated as random variables and the associated probability density function must be learned -- executing the optimization process is not trivial. Posterior sampling MCMC-based methods generally yield the most accurate posterior predictions \cite{papamarkou2022challenges}; however, their feasibility is limited by the dimensionality and complexity of the posterior distribution \cite{izmailov2021bayesian}. The class of VI techniques provides a comparatively scalable approximation in which the target posterior is replaced by a parameterized variational distribution. We explore both classes of methods to develop high-dimensional surrogate models for full-field predictions with uncertainty for materials applications in this work. 

Although there have been some research efforts geared towards predicting mean-field properties in mechanics of materials using ML and UQ algorithms, these have been either limited to relatively low dimensional effective properties \cite{olivier2021bayesian,jiang2023uncertainty}, to specific algorithm implementations \cite{zhu2018bayesian,mo2019deep,xia2022bayesian} or to the assessment of uncertainty in the model input parameters \cite{tripathy2018deep} rather than the model uncertainty. In fact, the application of UQ methods in NNs for high dimensional data is primarily found in segmentation tasks and in disparate disciplines such as computer vision \cite{kendall2017uncertainties}, biomedical segmentation \cite{hann2021ensemble,ng2018estimating} and earth imaging \cite{dechesne2021bayesian}. From the perspective of scientific applications, the comparative review of pertinent methods in \cite{yang2021b} has focused on fully-connected network architectures with a relatively small number of parameters aiming to solve well-posed PDEs. To the best of the authors' knowledge, there is no work focused on quantifying the uncertainty associated with full-field estimation using ML in solid mechanics applications. In an effort to address these gaps in the literature, the current study conducts a systematic assessment of pertinent CNN-based UQ methods with application to quantifying uncertainty in stress field predictions in stochastic material microstructures. 

This study is conducted by casting the prediction between the input microstructure and the output stress field as an image-to-image regression problem. First, a modified convolutional encoder-decoder NN architecture is employed as the surrogate model which captures the nonlinear relationship between the input and the output. Next, the parameters of the network are treated probabilistically and the posterior distribution is estimated using three state-of-the-art algorithms: the posterior-sampling-based Hamiltonian Monte Carlo (HMC) method, the variational Bayes by Backprop (BBB) algorithm, and the variational Monte Carlo Dropoout (MCD) technique. The efficacy of the proposed methodologies is demonstrated by considering two synthetic and statistically representative material datasets: a fiber reinforced composite and a polycrystalline material. The results are compared with FEA solutions and deterministic NNs, while the capability of the BNNs to provide interpretable uncertainty estimates in the predictions is investigated and benchmarked with respect to the state-of-the-art.

\section{Problem Statement}\label{section_2}
A persistent task in the field of computational mechanics relates to the prediction of material deformation and stress, which is conventionally provided by means of Finite Element Analysis (FEA) \cite{bathe2006finite,ghanem2003stochastic}. Factors such as complex microstructure topology and heterogeneity, and emerging nonlinear constitutive laws often require the solution of highly nonlinear partial differential equations \cite{reddy2015introduction,maurizi2022predicting}. This renders numerical simulations computationally prohibitive, especially when the goal relates to screening a large number of material microstructures. Addressing this shortcoming, NNs have shown promising results as cost-effective FEA surrogates in materials applications using a relatively limited amount of data \cite{bhaduri2022stress,maurizi2022predicting}. However, the challenge of estimating uncertainty in NN surrogate predictions remains. This is because the NN surrogate models are highly overparameterized and therefore UQ requires solving a very high-dimensional Bayesian inverse problem.  

\subsection{Bayesian neural network formulation}

The process of building a NN surrogate model for a high-dimensional FEA model entails optimizing the parameters (weights and/or biases) \(\boldsymbol{\omega} \in \mathbb{R}^d\) of a composite nonlinear function in the form \(f^{\boldsymbol{\omega}}(\mathbf{x})\) such that it approximates the relationship \(\left(f: \mathbf{x} \rightarrow \mathbf{y}\right)\) between the input \(\mathbf{x}\) and output \(\mathbf{y}\) training data, both of which can be given in the form of two-dimensional images. This is aimed at generating an accurate estimate \(\mathbf{y}^*\) for each new microstructure \(\mathbf{x}^*\), which is consistent with the training data distribution, through a forward pass of the network. In standard NNs, the parameters are typically learned by gradient-based minimization algorithms and they are represented by single point estimates. This conventional training methodology can also be viewed as an approximation to Bayesian minimization. Specifically, consider a set of training data \(\mathcal D = \{\mathbf{x}, \mathbf{y}\} = \{\mathbf{x}_i , \mathbf{y}_i\}_{i =1:n}\) where \(\mathbf{x}_i\) and \(\mathbf{y}_i\) denote the \(i^{th}\) input and output data sample, respectively. Adopting a Gaussian error assumption for the model predictions, the following probability model over the data is defined
\begin{equation}\label{likelihood}
        p\left(\mathbf{y} \mid \mathbf{x}, \mathcal{\boldsymbol{\omega}}\right)=\mathcal N (\mathbf{y}\mid \hat{f}(\mathbf{x}), \mathbf{\Sigma})
\end{equation}
where \(\mathcal N\) denotes the Gaussian distribution with mean \(\hat{f}(\mathbf{x})\) and covariance matrix \(\mathbf{\Sigma}\). This expression can be further simplified by assuming that the training data points are independent and identically distributed (i.i.d.). In this case, Eq.~\eqref{likelihood} takes the form 
\begin{equation}\label{likelihood_iid}
        p(\mathbf{y} \mid \mathbf{x}, \mathcal{\boldsymbol{\omega}}) = \prod_{i=1}^{n} \mathcal{N}(\mathbf{y}_i; \hat{f}_i(\mathbf{x}),\sigma^2_{\alpha})
\end{equation}
where \(\sigma^2_{\alpha}\) denotes the variance of the aleatoric uncertainty of the data. In a Bayesian framework, Eq.~\eqref{likelihood_iid} represents the likelihood of the data \(\mathcal D\) given the model parameters \(\mathbf{\boldsymbol{\omega}}\) and is also denoted by \(\mathcal{L}(\boldsymbol{\omega}|\mathcal{D}) = p(\mathcal{D}|\boldsymbol{\omega})\). The NN training task involves inferring the posterior distribution \(p(\boldsymbol{\omega}|\mathcal{D})\) of the weights given the observed data using Bayes' theorem in the form
\begin{equation}\label{Bayes}
    p(\mathbf{\boldsymbol{\omega}} \mid \mathcal D)=\frac{p(\mathcal D \mid \mathbf{\boldsymbol{\omega}}) p(\mathbf{\boldsymbol{\omega}})}{p(\mathcal D)}=\frac{p(\mathcal D, \mathbf{\boldsymbol{\omega}})}{\int_{\mathbf{\boldsymbol{\omega}}} p\left(\mathcal D, \mathbf{\boldsymbol{\omega}}^{\prime}\right) d \mathbf{\boldsymbol{\omega}}^{\prime}}
\end{equation}
where \(p(\boldsymbol{\omega})\) is the prior distribution, which encodes prior information or assumptions about the distribution of the weights before a data are observed. Further, \(p(\mathcal D)\) is the model evidence \cite{mackay1992bayesian}, which serves as a normalizing factor to ensure that the posterior distribution integrates to one (i.e. yields valid probabilities). 

In conventional NN training procedures the weights are learned by maximum likelihood estimation (MLE), whereby the likelihood $p(\mathcal D | \mathbf{\boldsymbol{\omega}})$ is maximized with respect to the model parameters, disregarding the prior distribution and the model evidence. Specifically, using the Gaussian assumption in Eq.~\eqref{likelihood_iid}, the logarithm of the likelihood function is expressed as \cite{bishop2006pattern}
\begin{equation}\label{log_likelihood}
    \log p(\mathcal{D} \mid \mathbf{\boldsymbol{\omega}})= -\frac{1}{2\sigma^2_{\alpha}} \sum_{i=1}^{n} \{\hat{\mathbf{y}}_i(\mathbf{x}|\boldsymbol{\omega}) - \mathbf{y}_i \}^2 - \frac{n}{2} \log \sigma^2_{\alpha} - \frac{n}{2} \ln(2\pi)
\end{equation}
where $\hat{\mathbf{y}}_i$ is the $i^{th}$ component of the NN prediction $\hat{\mathbf{y}}=\hat{f}(\mathbf{x})$. MLE is then performed by maximizing Eq.~\eqref{log_likelihood} with respect to \(\boldsymbol{\omega}\) as
\begin{equation}\label{MLE}
    \begin{aligned}
        \mathbf{\boldsymbol{\omega}}^{\mathrm{MLE}} & =\arg \max _{\mathbf{\boldsymbol{\omega}}} \log p(\mathcal{D} \mid \mathbf{\boldsymbol{\omega}}) \\ & =\arg \max _{\boldsymbol{\omega}} \sum_i \log p\left(\mathbf{y}_i \mid \mathbf{x}_i, \mathbf{\boldsymbol{\omega}}\right)
    \end{aligned}
\end{equation}
Clearly, considering Eq.~\eqref{log_likelihood} and \eqref{MLE}, maximizing the likelihood with respect to the parameters \(\boldsymbol{\omega}\) is equivalent to minimizing the mean square error (MSE) between the target values and the predictions. Determining \(\boldsymbol{\omega}\) using Eq.~\eqref{MLE} allows for making predictions, which are given in the form of point estimates.

In a similar manner, by considering that \(p(\mathbf{\boldsymbol{\omega}} \mid \mathcal D)\propto p(\mathcal D \mid \mathbf{\boldsymbol{\omega}}) p(\mathbf{\boldsymbol{\omega}})
\), it can also be shown that adding the prior distribution to the objective function in the form
\begin{equation}\label{MAP}
    \begin{aligned}
        \mathbf{\boldsymbol{\omega}}^{\mathrm{MAP}} & =\arg \max _{\mathbf{\boldsymbol{\omega}}} \log p(\mathbf{\boldsymbol{\omega}} \mid \mathcal{D}) \\ & =\arg \max _{\mathbf{\boldsymbol{\omega}}} \log p(\mathcal{D} \mid \mathbf{\boldsymbol{\omega}})+\log p(\mathbf{\boldsymbol{\omega}})
    \end{aligned}
\end{equation}
yields the maximum a posteriori (MAP) estimator, which identifies the most probable value of the posterior distribution of the weights.
Depending on the choice of prior distribution, this approach leads to distinct regularization schemes. For instance, adopting a Gaussian prior over the NN weights is equivalent to an \(l_2\)-norm penalization of the coefficients, also known as ridge regression \cite{zou2005regularization}. Similarly, employing a sparsity promoting Laplace prior leads to a \(l_1\)-norm penalization also known as basis pursuit denoising (BPDN) \cite{chen2001atomic} or least absolute shrinkage and selection operator (LASSO) \cite{tibshirani1996regression}. The aforementioned regularization techniques generally reduce model complexity and prevent overfitting. 

However, both minimization procedures yield deterministic estimates that account for the variance \(\sigma^2_{\alpha}\) (aleatoric uncertainty) but neglect uncertainty associated with the model parameters (epistemic uncertainty). This yields only a point estimate predictor and may lead to inconsistent predictions \cite{jospin2022hands}, {\it e.g.}, in cases where the posterior distribution is multimodal where the MAP estimate represents only a single mode of the distribution \cite{mitros2019validity}. Additionally, these point predictors may lead to unreliable predictions with respect to noise sensitivity and out-of-training-distribution data points \cite{mitros2019validity}.

To mitigate these limitations, a more comprehensive Bayesian inference formulation can be constructed by treating the NN weights as random variables and solving Eq.~\eqref{Bayes} directly to obtain $p(\mathbf{\boldsymbol{\omega}} | \mathcal D)$. In this setting, for each new input point \(\mathbf{x}^*\) the predictive probability of the output $\mathbf{y}^*$ can be found by marginalizing over the posterior weights as \cite{wilson2020bayesian}
\begin{equation}\label{predictive}
    p\left(\mathbf{y}^{\star} \mid \mathbf{x}^{\star}, \mathcal{\mathcal D}\right)=\int p\left(\mathbf{y}^{\star} \mid \mathbf{x}^{\star}, \mathbf{\boldsymbol{\omega}}\right) p(\mathbf{\boldsymbol{\omega}} \mid \mathcal D) d \mathbf{\boldsymbol{\omega}}
\end{equation}
having mean and variance
\begin{align}\label{predictive_mean}
    E[\mathbf{y}^* | \mathbf{x}^*, \mathcal{D}] &= E_{p(\boldsymbol{\omega}|\mathcal{D})} \left[ E[\mathbf{y}^* | \mathbf{x}^*, \boldsymbol{\omega}] \right] \\
    &= E_{p(\boldsymbol{\omega}|\mathcal{D})} \left[ f^{\boldsymbol{\omega}}(\mathbf{x}^*) \right] \nonumber
\end{align}
\begin{align}\label{predictive_variance}
    \text{Var}(\mathbf{y}^* | \mathbf{x}^*, \mathcal{D}) & = E_{p(\boldsymbol{\omega}|\mathcal{D})} \left[ \text{Var}(\mathbf{y}^* | \mathbf{x}^*, \boldsymbol{\omega}) \right] + \text{Var}_{p(\boldsymbol{\omega}|\mathcal{D})} \left( E[\mathbf{y}^* | \mathbf{x}^*, \boldsymbol{\omega}] \right)\\
    & = \sigma_{\alpha}^2(\mathbf{x}^*) + \text{Var}_{p(\boldsymbol{\omega}|\mathcal{D})} \left(f^{\boldsymbol{\omega}}(\mathbf{x}^*) \right)
\end{align}
where \(f^{\boldsymbol{\omega}}(\mathbf{x}^*)\) is the output of the NN for the set of parameters \(\boldsymbol{\omega}\) given input \(\mathbf{x}^{\star}\). Importantly, the first term in Eq.~\eqref{predictive_variance} is associated with noise in the data and corresponds to the error variance \(\sigma^2_{\alpha}\). This uncertainty is purely aleatory. Meanwhile, the second term is related to the uncertainty in the model parameters and corresponds to the variance of the posterior distribution of the weights \(p(\mathbf{\boldsymbol{\omega}} \mid \mathcal D)\). This uncertainty is epistemic and may be reduced by adding data. In this way, aleatory and epistemic uncertainty can be cleanly separated in the fully Bayesian NN.

\subsection{Neural network architecture for full-field material response}\label{NN_architecture}

The performance of NNs as surrogate models in full-field material response prediction ({\it e.g.,} as opposed to homogenized properties) depends on their ability to resolve spatially varying and localized features, such as areas of high stress concentration. For this reason, it is reasonable to treat the spatially varying field as an image. Although a plethora of architectures exist in the literature for various image processing tasks, convolutional neural networks (CNNs) are promising for their ability to extract hierarchical image features in high dimensional data. Specifically, encoder-decoder type convolutional architectures have been widely employed within the context of full-field estimation \cite{mo2019deep,bhaduri2022stress,khorrami2023artificial} due to their ability to implicitly obtain a low-dimensional representation of the input data and to efficiently propagate information through the encoding-decoding paths. It is noted that alternative suitable architectures exist, such as graph neural networks \cite{maurizi2022predicting} which do not rely on structured data. Nevertheless, an encoder-decoder CNN is adopted herein to account for the structured nature of the image data in our regression task. This choice is also motivated by the ubiquitous employment of CNNs in scientific applications, which prompts a deeper investigation into their UQ capabilities.

The NN employed in this study (illustrated in Figure \ref{Unet_BCNN_schematic}) is a modified version of the U-net convolutional architecture  which was originally introduced in \cite{ronneberger2015u} for biomedical image segmentation. The network comprises a series of encoding blocks and decoding blocks. Each encoding block consists of a repeated set of a convolutional layers of kernel size \(k\), a batch normalization layer followed by a nonlinear activation function and a downsampling layer. The commonly used rectified linear unit (ReLU) activation function is employed to introduce nonlinearity and to mitigate vanishing gradient issues. The maximum pooling (max-pooling) operation is used in the downsampling layer. These successive operations decrease the height and width of the original input by a factor of two at each step and increase the number of channels until the latent representation is reached. 

The decoding blocks have the same structure as their encoding counterparts with the exception that the downsampling layers are replaced by upsampling layers. In this manner, the inverse process is followed until the last convolutional layer of kernel size \(k = 1\) which combines the features of the last multi-channel output to a single prediction. The network also includes a number of skip connections between the contracting and expanding paths, aimed at combining high resolution features of the decoding block with abstract feature representations of the encoding block. This approach effectively propagates contextual information, accelerates convergence during training, and enhances the output resolution \cite{ronneberger2015u}.

A critical design parameter in this NN architecture is the number of filters at each encoding-decoding block. Typically, increasing the network depth allows the NN to capture complex features in the encoding space and enhances representation power, but it can lead to overfitting \cite{sun2016depth}. This strategy is further associated with increased computational overhead, especially when the NN is employed within a broader Bayesian framework. On this basis, the number of filters must be appropriately chosen to yield a trade-off between computational efficiency and accuracy.

Another key aspect that determines the NNs predictive performance is the choice of the kernel size, which defines the size of the matrix employed in each convolution operation. The underlying patterns are identified by optimizing the weights of the kernel matrix. It is expected that smoothly varying feature patterns can be captured accurately using a relatively small kernel size. For example, a small kernel size (\(k=3\)) has proven accurate for predicting the stress field of fiber-reinforced polymers \cite{sun2020predicting} and for predicting effective material properties in composite materials \cite{yang2018deep}. In contrast, a larger kernel size is more appropriate for highly heterogenous or anisotropic materials, in which sharp stress variations are anticipated. For instance, a relatively larger kernel size (\(k=9\)) \cite{mianroodi2021teaching, khorrami2023artificial} has been reported as a preferable value within the context of stress field prediction for polycrystalline materials. In view of the above, the kernel size is tailored to each numerical application. 

The U-net architecture presented above is implemented using the PyTorch ML library \cite{paszke2019pytorch} which enables graphics processing units (GPU) acceleration. Overall, the training procedure is described as follows. First, the parameters of the deterministic U-net are learned using the Adam \cite{kingma2014adam} gradient-based optimization algorithm. The loss function is defined as the MSE between the target and predicted output, which is a standard regression approach and corresponds to the MLE estimate described above. The learned parameter values are then utilized both for making deterministic predictions and for initializing fully Bayesian NN implementations to accelerate convergence and enhance accuracy. These Bayesian frameworks are presented in detail in the following section.

\subsection{Challenges in high-dimension}\label{Challenges}

In a probabilistic setting, obtaining the predictive posterior distribution in the form of Eq.~\eqref{predictive} requires performing Bayesian inference, which is a daunting task considering both the input data and the parameter space dimensionality. Specifically, calculating the integral in Eq.~\eqref{predictive} analytically, requires enumerating all possible combinations of the parameter values whose number increases combinatorially  with dimension. Additionally, the regions of probability concentration that dominate the expectation operation in Eq.~\eqref{predictive} tend to be localized in vanishingly small areas of the probability space as the dimension increases. Consequently, even a discrete approximation is challenging in high dimensions. Overall, modern NNs are characterized by a large number of parameters, e.g. hundreds of thousands or millions, which are learned using a comparatively much smaller amount of data (perhaps a few hundred or a few thousand). Therefore, Bayesian inference in NNs is a significantly ill-posed problem in which the posterior distribution is anticipated to be high-dimensional, multi-modal and non-convex \cite{izmailov2018averaging}. Despite the fact that it is not amenable to an analytical treatment, there exist several methodologies which allow  the predictive posterior distribution to be approximated in deep learning architectures. Some of the most commonly used approximations are described in the following section and employed in this work.

\section{Methodology}\label{Methodology}
In this study, the task of of data-driven full-field prediction is cast as a probabilistic image-to-image mapping using the NN architecture outlined in Section \ref{NN_architecture}. To achieve this, three Bayesian inference methods are employed to approximate Eqs.~\eqref{predictive}-\eqref{predictive_variance}: posterior sampling using Hamiltonian Monte Carlo (HMC), the Monte Carlo dropout (MCD) technique, and the variational Bayes By Backprop (BBB) algorithm. The first objective is to demonstrate the feasibility of HMC for UQ in field predictions for solid mechanics. To the best of our knowledge, this is the first time that HMC has been applied to image-to-image mappings for materials applications. The second objective, which also has not been addressed in the literature, is to systematically analyze the ability of these three UQ methods to quantify epistemic uncertainty in full-field material predictions.  HMC is regarded as the state-of-the-art inference method and theoretically yields the target posterior distribution; however, it requires significant computational effort. The more efficient MCD and BBB approaches are evaluated with respect to the HMC predictions. The following sections introduce these Bayesian algorithms and their implementation in the U-net architecture described above.

\subsection{Posterior Sampling using Hamiltonian Monte Carlo (HMC)}\label{HMC}
When the posterior distribution in Eq.~\eqref{predictive} is high-dimensional and/or complex, calculating the predictive distribution is intractable and approximate schemes are necessary. One way to estimate the posterior predictive distribution is to draw samples from it using the Markov chain Monte Carlo (MCMC) method \cite{gelfand1990sampling}, which constructs a Markov chain whose stationary distribution is the desired posterior distribution. Using the classical random walk MCMC, samples are drawn from a proposal distribution at each step of the chain and are accepted based on a Metropolis-Hastings (MH) acceptance criterion. Despite its wide application to engineering problems, random walk MH-based MCMC methods are not well-suited to high dimensional problems because, as the dimensionality of the distribution increases, the probability mass tends to be more concentrated in smaller regions of the probability space while the number of possible directions for generating samples increases exponentially \cite{betancourt2017conceptual}. This leads to an inefficient exploration of the probability space. Further, random-walk MH-based MCMC tends to yield highly correlated samples that are prone to become confined at local modes of the posterior distribution. Thus, a more effective strategy for exploring the probability space is required. 

The HMC method is a computationally efficient MCMC variant which mitigates the limitations of random walk MCMC \cite{yang2021b}. In HMC, the problem of sampling from the Markov chain \cite{neal2011mcmc} is performed by simulating the dynamics of an equivalent Hamiltonian system. Specifically, the random variables (weights of the NN) are viewed as position variables of the dynamical system and the original \(d\)-dimensional parameter space is augmented into a \(2d\)-dimensional phase space
\begin{equation}\label{hamiltonian_augment}
    \boldsymbol{\omega} \rightarrow (\boldsymbol{\omega}, \mathbf{r})
\end{equation}
where \(\mathbf{r}\) denotes an auxiliary momentum variable. The canonical distribution of the system is given by 
\begin{equation}\label{hamiltonian_canonical}
    p(\boldsymbol{\omega},\mathbf{r}) = \frac{1}{Z} \exp(-H(\boldsymbol{\omega},\mathbf{r}))
\end{equation}
where \(Z\) is a normalizing constant and \(H(\boldsymbol{\omega},\mathbf{r})\) is the
Hamiltonian energy function defined as 
\begin{equation}\label{hamiltonian_energy}
    H(\boldsymbol{\omega}, \mathbf{r}) = U(\boldsymbol{\omega}) + K(\mathbf{r})
\end{equation}
where \(U(\boldsymbol{\omega})\) and \(K(\mathbf{r})\) are the potential and kinetic energy functions, respectively. Applying Eq.~\eqref{hamiltonian_energy}, the distribution can be expressed in terms of the potential and kinetic energy terms as
\begin{equation}\label{hamiltonian_target_decomposed}
    p(\boldsymbol{\omega},\mathbf{r}) = p_\omega(\boldsymbol{\omega}) p_r(\mathbf{r})\propto \exp(-U(\boldsymbol{\omega}))\exp(-K(\mathbf{r})) 
\end{equation}
By definition, the kinetic energy function takes the following form
\begin{equation}\label{hamiltonian_kinetic}
    K(\mathbf{r}) = \frac{1}{2} \mathbf{r}^T \mathbf{M}^{-1} \mathbf{r}
\end{equation}
where the momentum variable is considered as Gaussian with a symmetric and positive-definite covariance matrix \(\mathbf{M}\). 
In the ensuing analysis, the standard HMC formulation is followed in which \(\mathbf{M}\) is further assumed to be a diagonal matrix.
By setting $p_\omega(\boldsymbol{\omega}) $ equal to the posterior distribution $p(\mathbf{\boldsymbol{\omega}} |\mathcal D)$ and applying Bayes' Rule from Eq.~\eqref{Bayes}, the potential energy \(U(\boldsymbol{\omega})\) can be expressed as
\begin{equation}\label{hamiltonian_potential}
    U(\boldsymbol{\omega}) \propto -\log \left[\mathcal{L}(\boldsymbol{\omega}|\mathcal{D}) p(\boldsymbol{\omega})\right]
\end{equation}
where \(\mathcal{L}(\boldsymbol{\omega}|\mathcal{D}) = p(\mathcal{D}|\boldsymbol{\omega})\) is the likelihood function. 

Next, concatenating the position and momentum variables into a \(2d\)-dimensional state vector \(\mathbf{z}=\{\boldsymbol{\omega},\mathbf{r}\}\), Hamilton's equations of motion are expressed by
\begin{equation}
    \frac{d\mathbf{z}}{dt} = \mathbf{J} \nabla H(\mathbf{z})
\end{equation}
where \(\nabla\) denotes the gradient operator, and \(\mathbf{J}\) is a \(2d \times 2d \) matrix defined as
\begin{equation}
    \mathbf{J} = \begin{bmatrix}
    \mathbf{0}_{d\times d} & \mathbf{I}_{d \times d} \\
    -\mathbf{I}_{d \times d} & \mathbf{0}_{d \times d}
    \end{bmatrix}
\end{equation}
where $\mathbf{0}_{d \times d}$ is the $d\times d$ matrix of zeros and $\mathbf{I}_{d \times d}$ is the $d\times d$ identity matrix. This formulation decouples Hamilton's equations into the following set of ordinary differential equations
\begin{align}\label{hamiltonian_ode}
    \begin{split}
        &\frac{d\omega_i}{dt} = \frac{\partial H}{\partial \omega_i}\\
        &\frac{dr_i}{dt} =  -\frac{\partial H}{\partial r_i} \ \ \ \forall \ i \ \in [1, \ldots, d].
    \end{split}
\end{align}    

The solution to Eq.~\eqref{hamiltonian_ode} is obtained numerically by employing symplectic integrators \cite{blanes2014numerical}, which satisfy the following conditions. First, they are time reversible and therefore generate samples that are consistent with reversible Markov transitions \cite{neal2011mcmc}. Second, they preserve the phase space volume in accordance with the Hamiltonian dynamics and, thus, they limit the error accumulation over multiple iterations. In this paper, the commonly used Stromer-Verlet (leapfrog) integrator is employed. The dynamics are numerically simulated using the leapfrog stepsize \(\varepsilon\) for \(L\) number of steps, thus defining the total simulation time per step as \(T = L \varepsilon\). The interested reader is referred to \cite{girolami2011riemann} for further details.

The HMC algorithm generates samples from the target probability distribution which takes the form of the canonical distribution in Eq.~\eqref{hamiltonian_canonical}. In the standard implementation, sample generation consists of two steps. First, starting from an initial position \(\boldsymbol{\omega}_j\) the momentum variable \(\mathbf{r}_j\) is sampled randomly from a zero mean, multivariate Gaussian distribution \(p(\mathbf{r}) \sim \mathcal{N}(0,\mathbf{M})\). Next, the Hamiltonian dynamics are simulated using the symplectic integrator for a time $T$ to generate a proposal state \((\boldsymbol{\omega}^*,\mathbf{r}^*)\). This proposal state is accepted with probability \(\alpha\) given by the following Metropolis-Hastings acceptance criterion
\begin{equation}\label{hamiltonian_Metropolis}
    \alpha = \min \left[ 1, \exp\left(-H(\boldsymbol{\omega}^*, p^*) + H(\boldsymbol{\omega}_j, \mathbf{r}_j)\right) \right]
\end{equation}

The HMC algorithm is theoretically guaranteed to generate samples that asymptotically converge to the true posterior. However, with a few exceptions (e.g. \cite{yang2021deep}) its application has been limited to relatively low-dimensional problems (e.g., classification \cite{cobb2021scaling}) because it becomes computationally prohibitive with increasing numbers of data points and model parameters \cite{kass1995bayes}. Further, it heavily relies on the optimal selection of the stepsize \(\varepsilon\) and the number of leapfrog steps \(L\), which are problem dependent and require trial and error tuning. It has been observed that, as the complexity of the model increases, a higher trajectory length is preferred to ensure a systematic exploration of the state space \cite{neal2011mcmc}. Clearly, this imposes a significant computational burden. One adaptive scheme, known as the No-U-Turn Sampler (NUTS) \cite{hoffman2014no}, allows the time integration parameters to be automatically determined, but it requires an arduous parameter identification process. Another computationally enhanced approach relies on data mini-batching \cite{welling2011bayesian}, which however, leads to perturbation of the stationary distribution \cite{izmailov2021bayesian}. Lastly, a recently proposed full-batch approach based on the symmetric splitting of the Hamiltonian renders HMC amenable to convolutional NNs. However, it considers short trajectory lengths (\(L\leq 100\)) and relatively small-scale models. In this work, we demonstrate that HMC can be applied for modestly large CNN architectures for material field predictions. 

\subsection{Variational inference}\label{VI}
An efficient alternative method for evaluating the integral in Eq.~\eqref{predictive} is to apply variational inference (VI). In this setting, the exact posterior is substituted by a parametric variational posterior distribution \(q_{\boldsymbol{\theta}}(\boldsymbol{\omega})\) having parameters \(\boldsymbol{\theta}\). Several VI approaches exist, 
such as the Bayes by Backprop algorithm, the \(\alpha\)-Blackbox divergence minimization, methods that utilize stochastic gradients of the variational objective function \cite{kingma2013auto,ranganath2014black}, variational inference coupled with latent variable models \cite{rezende2015variational}, etc. Ultimately, the majority of these methods require stochastic gradient evaluations of the evidence lower bound (ELBO). Typical examples in scientific applications include the application of the Stein Variational Gradient Descent (SVGD) algorithm for the estimation of two-dimensional random permeability fields \cite{zhu2018bayesian} and the use of variational autoencoders (VAEs) for performing multiscale analysis \cite{xia2022bayesian}, medical segmentation and urban scene understanding \mbox{\cite{monteiro2020stochastic, kohl2018probabilistic,kohl2019hierarchical}}. 

In any VI approach, the variational parameters are found by minimizing the discrepancy between the true posterior and the variational distribution. Various discrepancy measures have been proposed in the literature including methods based on the alpha divergence~\cite{hernandez2016black} and Jensen-Shannon divergence~\cite{thiagarajan2022jensen}, among others. The most commonly employed, and the method chosen in this study, employs the Kullback–Leibler (KL) divergence such that the minimization is cast as follows:
\begin{equation}\label{KL}
    \begin{aligned} \boldsymbol{\theta}^{\star} & =\arg \min _{\boldsymbol{\theta}} \mathrm{KL}[q_{\boldsymbol{\theta}}(\boldsymbol{\omega}) \| p(\mathbf{\boldsymbol{\omega}} \mid \mathcal{D})].\end{aligned}
\end{equation}
Applying Bayes' theorem in Eq.~\eqref{Bayes} for \(p(\mathbf{\boldsymbol{\omega}} | \mathcal{D})\) and considering that \(\mathrm{KL}[q \| p] = \mathbb{E}_q \left[ \ln\left(\frac{q}{p}\right) \right]\), the minimization procedure to find the optimal variational parameters \(\boldsymbol{\theta}\) becomes independent from the log model evidence calculation. Thus, the optimal parameters \(\boldsymbol{\theta}^*\) are equivalently found by
\begin{equation}\label{VFE}
    \boldsymbol{\theta}^{\star} =  \arg \min _{\boldsymbol{\theta}}  -\mathbb{E}_{q_{\boldsymbol{\theta}}(\boldsymbol{\omega})}[\log p(\mathcal{D}|\boldsymbol{\omega})] + \lambda \cdot KL[q_{\boldsymbol{\theta}}(\boldsymbol{\omega}) \| p(\boldsymbol{\omega})]    
\end{equation}
which denotes minimizing the variational free energy. The expression on the right hand side of Eq.~\eqref{VFE} is the upper bound of the KL divergence. Minimizing this bound effectively minimizes the divergence between the variational posterior and the true posterior \cite{millidge2021whence}. Thus, the objective function in Eq.~\eqref{VFE} is also called the evidence lower bound (ELBO). The first term is the expected log-likelihood of the data given the variational parameters \(\boldsymbol{\theta}\) and the second term is the KL divergence between the variational posterior and the prior distribution. This can be interpreted as a decomposition between model accuracy and model complexity, respectively in which the regularization parameter \(\lambda\) controls the trade-off between the two terms.  

In this work, two widely employed mean field VI approaches in deep learning are applied: the MCD and BBB algorithms. These methods are reviewed next.

\subsubsection{Monte Carlo dropout (MCD)}\label{MCD}
The MCD technique has been widely employed as a regularization method  by removing nodes from a number of layers during training with a specified probability rate, e.g., Bernoulli or Gaussian \cite{srivastava2014dropout}. It has also been shown that applying dropout during prediction yields a distribution of outputs, thus, providing an approximation of model uncertainty \cite{gal2016dropout}. 
To achieve this, the NN is cast into a deep Gaussian process (GP) with \(L\) layers, and the weights are used to parameterize the covariance function at each layer by
\begin{equation}\label{dropout_weights}
    \mathbf{W}_i =\mathbf{\mathcal{M}}_i \cdot \operatorname{diag}\left(\left[{z}_{i, j}\right]_{j=1}^{K_i}\right), 
\end{equation}
\begin{equation}\label{dropout_bernoulli}
    {z}_{i, j} \sim \operatorname{Bernoulli}\left(p_i\right) \text { for } i=1, \ldots, L, j=1, \ldots, K_{i-1}
\end{equation}
where \(i\) denotes the layer index, \(\mathbf{W}_i\) and \(\mathcal{M}_i\) are the matrix of weights for each layer before and after dropout, respectively, \(K_i\) denotes the size of each layer and matrix, and \(p_i\) is the dropout probability. This formulation yields the variational posterior as the distribution over matrices whose columns are set to zero defined as
\begin{equation}
    q(\boldsymbol{\omega}) = \prod_{i=1}^{L} q(\mathbf{W}_i  \mid \boldsymbol{\theta}_i)
\end{equation} 
in which the network parameters are given as random variables in the form \(\boldsymbol{\omega}=\left\{\mathbf{W}_i\right\}_{i=1}^L\), and the variational parameters at each layer are given by \(\boldsymbol{\theta}_i = \left\{M_i, p_i\right\}\).

The predictive mean is empirically calculated by sampling \(N_s\) realizations of the set of Bernoulli vectors \(\mathbf{z}_i^t=\left[{z^t}_{i, j}\right]_{j=1}^{K_i}, t=1,\dots, N_s\), yielding \(N_s\) sets of network parameters \(\left\{\mathbf{W}_1^t, \ldots, \mathbf{W}_L^t \right\}\). This amounts to performing \(N_s\) forward passes of the network where, at each run, the weights are sampled from the underlying Bernoulli distribution to yield a prediction \(\widehat{\mathbf{y}}^*\) for each new input \(\mathbf{x}^*\). Following Eq.~\eqref{predictive_mean}, the output mean is then calculated by averaging the ensemble predictions in the form 
\begin{equation}\label{MCD_mean}
    \mathbb{E}_{q\left(\mathbf{y}^* \mid \mathbf{x}^*\right)}\left(\mathbf{y}^*\right) \approx \frac{1}{N_s} \sum_{t=1}^T \widehat{\mathbf{y}}^*\left(\mathbf{x}^*, \mathbf{W}_1^t, \ldots, \mathbf{W}_L^t\right).
\end{equation}
and the prediction uncertainty is approximated by estimating the variance from Eq.~\eqref{predictive_variance} by,
\begin{equation}\label{MCD_variance}
    \text{Var}_{q(\mathbf{y}^*|\mathbf{x}^*)} (\mathbf{y}^*) \approx \sigma_{\alpha}^2\mathbf{I}_n + \frac{1}{N_s} \sum_{t=1}^T \mathbf{y}^*(\mathbf{x}^*, \mathbf{W}_t)^T \mathbf{y}^*(\mathbf{x}^*, \mathbf{W}_t) - \mathbb{E}_{q(\mathbf{y}^*|\mathbf{x}^*)}(\mathbf{y}^*)^T \mathbb{E}_{q(\mathbf{y}^*|\mathbf{x}^*)}(\mathbf{y}^*)
\end{equation}
where \(\sigma_{\alpha}^2\) is the variance of the likelihood defined in Eq.~\eqref{likelihood_iid}. The prediction variance is then calculated using the sample output variance and adding the assumed aleatoric variance.

The MCD technique can be readily incorporated into standard NN architectures, which makes it amenable to high-dimensional problems.  For instance, in \cite{kendall2015bayesian} a CNN endowed with dropout layers was used for pixel-wise image classification with uncertainty and dropout layers have been used in U-net architectures for earth observation image classification \cite{dechesne2021bayesian}, three-dimensional material imaging \cite{labonte2019we} and biomedical segmentation \cite{hann2021ensemble}. This makes it relatively straightforward for our application of full-field material response prediction. However, the approximated posterior is highly dependent on the dropout rate and the number of dropout layers \cite{verdoja2020notes}, while additional data do not necessarily lead to posterior convergence. This makes training by MCD highly heuristic. In fact, it has been argued that MCD approximates the risk of a fixed model rather than estimating model uncertainty \cite{osband2016risk}. Further, the method yields overconfident predictions with respect to data covariance shift \cite{chan2020unlabelled}. 

\subsubsection{Bayes by Backprop (BBB)}\label{BBB}
An alternative methodology for performing UQ combines VI with the backpropagation algorithm used for training standard neural networks. This leads to the Bayes By Backprop alogrithm \cite{blundell2015weight} in which the ELBO is numerically approximated by Monte Carlo integration, yielding the following minimization (from Eq.~\eqref{VFE}): 
\begin{equation}\label{BBB_VFE}
    \boldsymbol{\theta}^* = \arg \min_{\boldsymbol{\theta}}  \sum_{i=1}^{M} \left[ - \log p(\mathcal{D}|\boldsymbol{\omega}^{(i)}) + \lambda \left( \log q_{\boldsymbol{\theta}}(\boldsymbol{\omega}^{(i)}) - \log p(\boldsymbol{\omega}^{(i)}) \right) \right]
\end{equation}
where \(M\) is the number of Monte Carlo samples. 
In BBB, the posterior weights are modeled by \(\boldsymbol{\mu}+ \log(1+\exp(\boldsymbol{\rho})) \circ \boldsymbol{\epsilon}\) in which \(\circ\) denotes the element-wise product. Here, $\boldsymbol{\epsilon}$ is a vector of uncorrelated standard normal random variables, the unknown (optimized/learned) vector  \(\boldsymbol{\theta} = (\boldsymbol{\mu},\boldsymbol{\rho})\) is comprised of the mean values \(\boldsymbol{\mu} = (\mu_1,...,\mu_{d})\) and standard deviations \(\boldsymbol{\rho} = (\rho_1,...,\rho_{d})\) of the individual Gaussian weights in the network, and \(\log(1+\exp(\cdot))\) is the softplus function, which ensures that the standard deviation is always positive. This formulation defines an i.i.d.\ random vector $\boldsymbol{\epsilon}$ such that each network parameter can be sampled independently in a computationally efficient manner. Additionally, derivatives of the objective function in Eq.~\eqref{BBB_VFE} can be evaluated with respect to the variational parameters to enable standard gradient-based optimization algorithms, which is essential for backpropagation. This approach is also known as the reparameterization trick \cite{kingma2013auto}.

In the standard form of the BBB algorithm employed here, a fully factorized Gaussian prior is adopted with mean \(\mu_{prior}\) and standard deviation \(\sigma_{prior}\) in the form \(\mathcal{N}(\mu_{prior}\mathbf{I}, \sigma_{prior}\mathbf{I})\). The algorithm requires as inputs the set of input-output data \(\mathcal D = \{\mathbf{x}, \mathbf{y}\}\), the number of Monte Carlo samples \( M \) to be drawn from the variational posterior, the number of training epochs \( E \) and the values for the hyperparameters \( \lambda, \alpha_{\boldsymbol{\mu}}, \alpha_{\boldsymbol{\rho}} \). The algorithm starts by initializing the variational posterior parameters \(\boldsymbol{\theta}_{\text{prior}} = (\mu_{\text{prior}}, \sigma_{\text{prior}})\) and at each training epoch \( j = 1, 2, \ldots, E \), the following steps are performed:

The noise variable is drawn independently from a zero mean Gaussian distribution \(M\) times
\begin{equation}\label{BBB_step1}
  \boldsymbol{\epsilon}^{(k)} \sim \mathcal{N}(0, \mathbf{I}_{d_{\boldsymbol{\theta}}}), \quad \text{for} \, k = 1, 2, \ldots, M
\end{equation}
The noise samples are then used to calculate the posterior weights in the form
\begin{equation}\label{BBB_step2}
    \boldsymbol{\omega}^{(k)} = \boldsymbol{\mu} + \log(1 + \exp(\boldsymbol{\rho})) \odot \boldsymbol{\epsilon}^{(k)}, \, \text{for} \, k = 1, 2, \ldots, M
\end{equation}
which are used to compute the loss function given by 
\begin{equation}\label{BBB_step3}
    L(\boldsymbol{\theta}) = \frac{1}{M}\sum_{i=1}^{M} \left[ - \log p(\mathcal{D}|\boldsymbol{\omega}^{(i)}) + \lambda \left( \log q_{\boldsymbol{\theta}}(\boldsymbol{\omega}^{(i)}) - \log p(\boldsymbol{\omega}^{(i)}) \right) \right]
\end{equation}
Next, the gradients are with respect to the variational parameters are calculated using the Backpropagation algorithm
\begin{equation}\label{BBB_step4}
   \Delta \boldsymbol{\mu} = \frac{\partial L(\boldsymbol{\theta})}{\partial \boldsymbol{\mu}}, \quad \Delta \boldsymbol{\rho} = \frac{\partial L(\boldsymbol{\theta})}{\partial \boldsymbol{\rho}}
\end{equation}
Finally, the parameters are updated based on the gradients in the form 
\begin{align}
    \boldsymbol{\mu} &\leftarrow \boldsymbol{\mu} - \alpha_{\boldsymbol{\mu}} \Delta \boldsymbol{\mu} \label{BBB_step4a}\\
    \boldsymbol{\rho} &\leftarrow \boldsymbol{\rho} - \alpha_{\boldsymbol{\rho}} \Delta \boldsymbol{\rho} \label{BBB_step4b}
\end{align}

Once the training process is complete, the predictive mean and variance given a new input point \(\mathbf{x}^*\) are calculated using Monte Carlo estimates for Eqs.~\eqref{predictive_mean} and \eqref{predictive_variance}. This is achieved by sampling the parameters of the variational posterior \(N_s\) times and evaluating
\begin{align}\label{BBB_mean}
    \mathbb{E}_{p(\mathbf{y}^*|\mathbf{x}^*, D)}[\mathbf{y}^*] & = \int  p(\mathbf{y}^*|\mathbf{x}^*, \mathbf{w}) q_{\theta}(\mathbf{w}|D) d\mathbf{w}\\
    & \approx \frac{1}{N_s} \sum_{t=1}^T p(\mathbf{y}^*|\mathbf{x}^*, \mathbf{w}_t) \nonumber
\end{align}
\begin{equation}\label{BBB_variance}
    \text{Var}_{q(\mathbf{y}^*|\mathbf{x}^*)}(\mathbf{y}^*) = \mathbb{E}_q[\mathbf{y}^*\mathbf{y}^{*T}] - \mathbb{E}_q[\mathbf{y}^*]\mathbb{E}_q[\mathbf{y}^*]^T
\end{equation}

The BBB algorithm exhibits good computational performance. The use of independent Gaussian distributions for each parameter leads to a reasonable number of optimization variables and allows sampling of those parameters at each optimization step. It can also be readily applied with existing stochastic gradient descent optimization procedures.  The resulting computational scalability allows for its application in a high-dimensional setting. For instance, BBB has been used to predict effective material properties from high dimensional microstructure images \cite{olivier2021bayesian}, to segment magnetic resonance images \cite{ng2018estimating} and to perform pixel-level segmentation of corrosion \cite{Nash_Zheng_Birbilis_2022}.

\subsection{Bayesian neural network training}
\begin{figure}
    \centering
    \includegraphics[width=\textwidth]{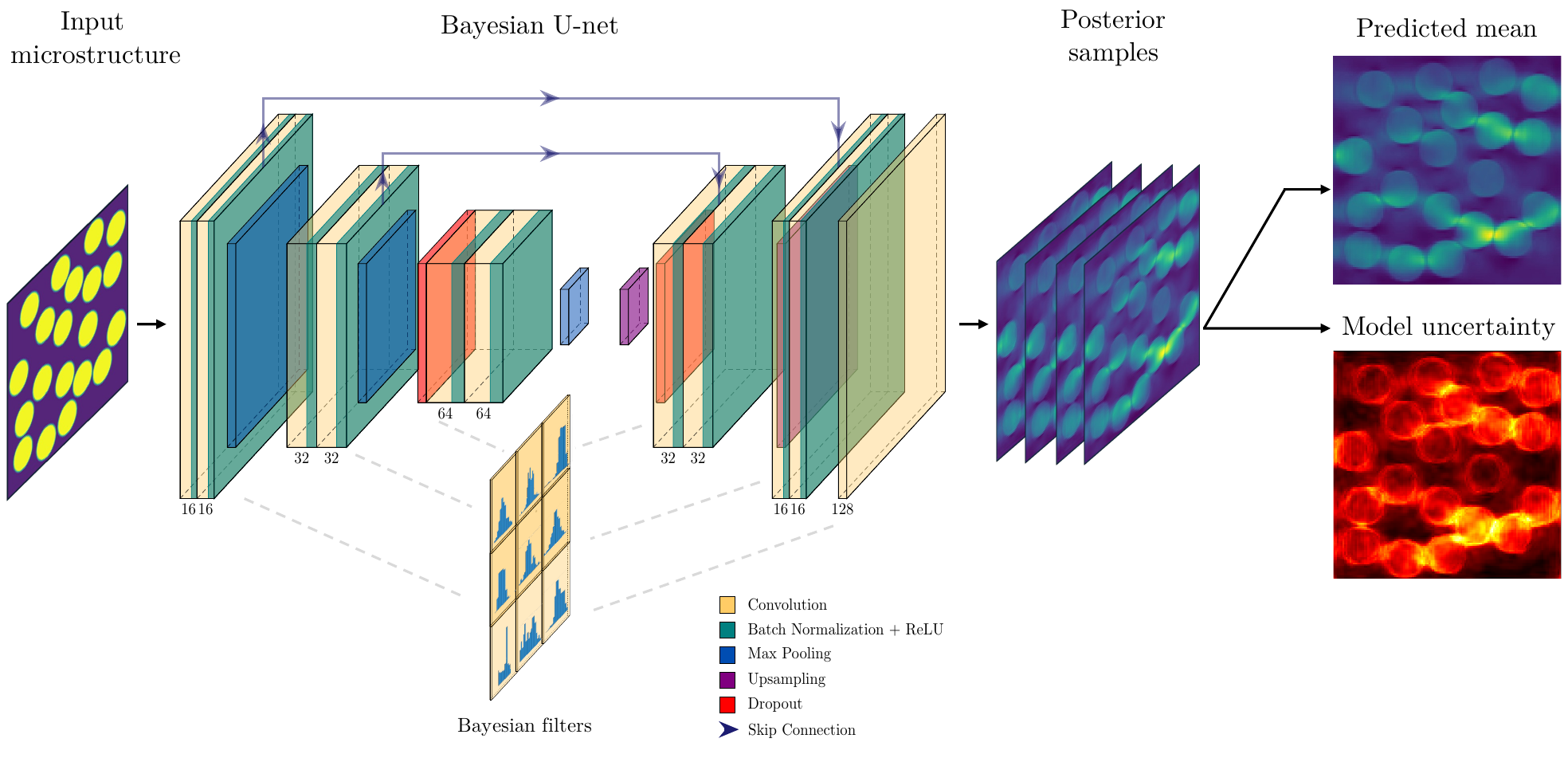}
    \caption{Schematic representation of the surrogate modeling and uncertainty quantification scheme. The U-net architecture incorporates dropout layers for the MCD technique or probabilistic filter parameters for the BBB and HMC methods. For predictions, the dropout layers or the Bayesian parameters are sampled to generate multiple stress predictions. The mean prediction is obtained averaging the resulting fields and model uncertainty is estimated by computing the standard deviation of the ensemble predictions.}
    \label{Unet_BCNN_schematic}
\end{figure}

Recall that our objective is to systematically compare the three Bayesian methodologies for neural network training in image-to-image regression tasks for materials applications, in terms of computational feasibility/tractability and their representations of \textit{epistemic} uncertainty in predicting full-field material response (i.e. prediction uncertainty at each pixel). To achieve this systematic comparison, we require a consistent training scheme on a common network architecture across all three algorithms. This is shown schematically in Fig.~\ref{Unet_BCNN_schematic}.  

In all cases, the likelihood function is assumed to be Gaussian as \(\mathcal{L}(\mathcal{D}|\boldsymbol{\omega}) \sim \mathcal{N}( g(\mathbf{x},\boldsymbol{\omega}),\sigma_{\alpha}^2\mathbf{I})\) where \(g(\mathbf{x},\boldsymbol{\omega}) = \frac{1}{n} \sum_{i=1}^n (\boldsymbol{y}_i - f^{\boldsymbol{\omega}}(\mathbf{x}_i))^2\), and the prior distribution is similarly assumed Gaussian  \(p(\boldsymbol{\omega}) \sim \mathcal{N}(\mathbf{m}_{\mathbf{\omega}},\sigma_{prior}^2 \mathbf{I})\). The term \(\mathbf{m}_{\mathbf{\omega}}\) denotes the prior mean and the terms \(\sigma_{\alpha}^2\) and \(\sigma_{prior}^2\) denote the likelihood and prior variances, respectively. The term \(\sigma_{\alpha}^2\) typically quantifies the degree of aleatoric uncertainty, which may be known, assumed, or estimated from the data. In the present study, each microstructure has a unique representation in the input-output space and noise is exclusively introduced by pixel discretization errors, which are considered very small. Therefore, \(\sigma_{\alpha}^2\) is expected to be very low and our primary objective is to estimate \textit{epistemic} uncertainty associated with the neural network predictions for a given (nearly noise free) microstructural realizations. Meanwhile, the prior variance \(\sigma_{prior}^2\) is set to balance between model accuracy and model complexity \cite{izmailov2021bayesian,cobb2021scaling}. 

While the overall architecture (see \hyperref[Appendix]{Appendix}) and probabilistic setting are set to be as consistent as possible, there are, nonetheless, aspects of each Bayesian implementation that are unique. For HMC, a deterministic pre-training is performed. The potential energy function \(U(\boldsymbol{\omega})\) in Eq.~\eqref{hamiltonian_potential} is defined using the prior and likelihood assumptions mentioned above where \(\mathbf{m}_{\mathbf{\omega}}\) is set equal to the parameters obtained from deterministic training and a relatively small \(\sigma_{prior}^2\) is chosen to reflect confidence in the pre-trained model, but also to yield adequate prediction accuracy. In addition, the burn-in period is set to zero since it is assumed that the deterministic training parameters are drawn from the stationary distribution. Following the HMC tuning strategy proposed by Neal et al.~\cite{neal2011mcmc} for high-dimensional problems, the step-size \(\epsilon\) is reduced until a high acceptance probability is achieved and the trajectory length \(L\) is increased to jointly ensure systematic exploration of the state space and low inter-sample correlation. Exact parameter values are provided along with the results in Section \ref{Results}.

The MCD framework is implemented by placing several dropout layers with a specified dropout probability \(p\) at distinct locations in the encoding and decoding paths of the U-net architecture. Using the same value for \(\sigma_{\alpha}^2\) as in the HMC method and employing the learned parameters from the deterministic model, variational inference is performed by computing \(N_s\) forward passes of the network for each new set of input data \(\mathbf{x}^*\). The predictive mean and variance are then calculated using Eq.~\eqref{MCD_mean} and ~\eqref{MCD_variance}.

Finally, the BBB algorithm is applied by assuming that the weights in the convolutional layers are Gaussian random variables in the original U-net architecture. The implementation of the proposed scheme utilizes the framework developed by Shridhar and coworkers \cite{shridhar2019comprehensive}, which incorporates Bayesian layers into CNNs in a straightforward manner. The mean and the variance of each parameter is trained using the BBB algorithm in Section \ref{BBB} with the same number of epochs \(E\) as in the deterministic case and drawing \(M=1\) Monte Carlo samples per iteration. Also, the likelihood function and the value for \(\sigma_{\alpha}^2\) are the same as those used in the HMC method. On the other hand, a zero mean prior is adopted for all weights \(\mathbf{m}_{\mathbf{\omega}} = \mathbf{0}\) and deterministic pre-training is not performed. The learning rates of \( \alpha_{\boldsymbol{\mu}}\) and \(\alpha_{\boldsymbol{\rho}}\) are set based on the Adam optimization learning rate while the value for \(\lambda\), which controls the variational penalty term, is chosen such that the terms in Eq.~\eqref{VFE} are have the same order of magnitude. The predictive mean and variance for each new input dataset \(\mathbf{x}^*\) are calculated using Eqs.~\eqref{BBB_mean} and ~\eqref{BBB_variance}, by sampling the network weight \(N_s\) times.

\section{Results and discussion}\label{Results}

In this section, we apply the three Bayesian inference schemes described in Section \ref{Methodology} for UQ in stress field prediction for two model materials: a heterogeneous fiber reinforced composite microstructure and a polycrystalline microstructure. We first describe the data generation process and establish the metrics used for comparison. We then present the results for each material. 

\subsection{Data generation and metrics for comparison}\label{Data_and_metrics}
The input data for both case studies are two dimensional synthetic microstructures given as images represented on a square pixel grid. For each input microstructure the associated stress field (the output data) is generated by solving a linear elasticity problem using the finite element method (FEM) under the same boundary conditions across all samples. The resulting stress field is interpolated to conform to the same pixel grid discretization as the input. Both input-output data pairs have dimensions \(n\times N_C\times N_H \times N_W\) where \(n=1140\) is the number of samples, \(N_C = 1\) is the number of channels, \(N_H = 128\) pixels is the height and \(N_W= 128\) pixels is the width of the images. The data is then partitioned into different subsets, with 1000 images designated for training (\(n_{\text{train}} = 1000\)), 100 images for validation (\(n_{\text{validation}} = 100\)), and 40 images for testing (\(n_{\text{test}} = 40\)). Note that it is possible to obtain multiple field estimates ({\it e.g.}, the components of the stress tensor) for each unique input microstructure, when the output is cast into a multi-channel NN as previously demonstrated \cite{bhaduri2022stress,nie2020stress}. However, without loss of generality, this study accounts only for a single field output, specifically the \(\sigma_{11}\) stress component, for the purpose of simplicity in presentation. 

A systematic assessment is performed by considering the predictions for each sample of the posterior \(\hat{\mathbf{y}}_{ij} \in \mathbb{R}^{N_H\times N_W}, \ i \in [1, \ldots, n], \ j \in [1, \ldots, N_s]\) to compute the following statistical estimates
\begin{align}
    \hat{\boldsymbol{\mu}}_i & = \frac{1}{N_s} \sum_{j=1}^{N_s} \hat{\mathbf{y}}_{ij} \label{mu_prediction} \\
    \hat{\boldsymbol{\sigma}}_i & = \sqrt{\frac{1}{N_s} \sum_{j=1}^{N_s} (\hat{\mathbf{y}}_{ij} - \hat{\boldsymbol{\mu}}_i)^2} \label{s_prediction}
\end{align}    
where \(\hat{\boldsymbol{\mu}}_i\) and \(\hat{\boldsymbol{\sigma}}_i\) are of dimensions \({N_H\times N_W}\) and denote the mean and the standard deviation of the predicted stress for the \(i^{\text{th}}\) microstructure, respectively. The scalar \textit{average standard deviation} across all samples in the test dataset is calculated by
\begin{equation}\label{s_avg}
    \sigma_{avg} = \sqrt{\frac{1}{N_{\text{test}} N_H N _W} \sum_{i=1}^{N_{\text{test}}} \sum_{j=1}^{N_H}\sum_{k=1}^{N_W} \hat{\boldsymbol{\sigma}}_{ijk}},
\end{equation}
which gives a single scalar measure of overall variability. Further, using Eqs.~\eqref{mu_prediction} and the FE-based stress fields \(\mathbf{y}_i, \in \mathbb{R}^{N_H\times N_W}, \ i \in [1, \ldots, n]\), the \textit{absolute error} is given by 
\begin{equation}\label{AE}
    \boldsymbol{\varepsilon}_{\hat{\boldsymbol{\mu}}_i} = \left| \mathbf{y}_{i} - \hat{\boldsymbol{\mu}}_{i} \right|
\end{equation}
This allows calculation of the \textit{mean absolute prediction error} and the associated \textit{prediction error standard deviation} across all samples and pixels in the dataset:
\begin{align}
    \mu_{\boldsymbol{\varepsilon}} &= \frac{1}{n N_H N_W} \sum_{i=1}^{n} \sum_{j=1}^{N_H} \sum_{k=1}^{N_W} \boldsymbol{\varepsilon}_{\hat{\boldsymbol{\mu}}_{ijk}} \label{mae}\\
    \sigma_{\boldsymbol{\varepsilon}} &= \sqrt{\frac{1}{n N_H N_W} \sum_{i=1}^{n}   \sum_{j=1}^{N_H} \sum_{k=1}^{N_W} \left(\boldsymbol{\varepsilon}_{ijk} -  \mu_{\boldsymbol{\varepsilon}}^2\right)}\label{error_variance}
\end{align}
These provide scalar measures of the overall accuracy and the variability in accuracy across the sample set.
Finally, considering the HMC as a reference UQ methodology, the error of the prediction variance relative to the HMC-based variance, referred to as \textit{HMC-relative variance error}, is given as
\begin{equation}\label{HMC_rel_var}
    \boldsymbol{\varepsilon}_{\hat{\boldsymbol{\sigma}}_i} = \left| \boldsymbol{\sigma}_{\mathrm{HMC}_i} - \hat{\boldsymbol{\sigma}}_i \right|
\end{equation}
Similarly, the \textit{mean HMC-relative variance error} and \textit{HMC-relative variance error standard deviation} are given by
\begin{equation}\label{variance_mae}
    \mu_{\boldsymbol{\varepsilon}_{\hat{\boldsymbol{\sigma}}}} = \frac{1}{n N_H N_W} \sum_{i=1}^{n} \sum_{j=1}^{N_H} \sum_{k=1}^{N_W} \boldsymbol{\varepsilon}_{\hat{\boldsymbol{\sigma}}_{ijk}}
\end{equation}
\begin{equation}\label{variance_error_variance}
    \sigma_{\boldsymbol{\varepsilon}_{\hat{\boldsymbol{\sigma}}}} = \sqrt{\frac{1}{n N_H N_W} \sum_{i=1}^{n} \sum_{j=1}^{N_H} \sum_{k=1}^{N_W} \left( \boldsymbol{\varepsilon}_{\hat{\boldsymbol{\sigma}}_{ijk}} - \mu_{\boldsymbol{\varepsilon}_{\hat{\boldsymbol{\sigma}}}} \right)^2}
\end{equation}
These are used to provide a spatially varying measure of variability for BBB and MCD relative to HMC (Eq.~\eqref{HMC_rel_var}) and overall measures of relative variability (Eqs.~\eqref{variance_mae}--\eqref{variance_error_variance}).

\subsection{Fiber-reinforced composite}
In the first case study, a two-dimensional cross section of a fiber-reinforced composite under plane strain is examined. The material behavior is assumed to be linearly elastic, and the fiber and matrix constituents are assumed to be perfectly bonded. The Young's moduli are \(E_{fiber} = 87 \ \textrm{GPa}\) and \(E_{matrix} = 3.2 \ \textrm{GPa}\), and the Poisson's ratios are \(\nu_{fiber} = 0.2\) and \(\nu_{matrix} = 0.35\). The number of fibers $N_{fiber}=20$, the fiber radius $r_{fiber}=0.6383$ and the volume fraction $v_f=0.4$ are held constant for each microstructure and \(285\) microstructures with random spatial fiber distributions are generated. Next, the spatially varying stress fields in each sample are calculated by applying a horizontal tensile displacement with magnitude \(0.008\) (corresponding to 0.1\% applied strain with traction-free boundary conditions on the top and bottom surfaces, using the ABAQUS \cite{standard1995user} software suite). As previously noted, the $\sigma_{11}(\bm{x})$ component of the stress tensor is retained to define the output as a single channel image. The total number of images is increased to \(n=1140\) by exploiting two planes of symmetry along the center-lines in the x and  y directions for data augmentation as detailed in \cite{bhaduri2022stress}. The mesh size for the finite element analyses is very fine, having $\approx$86000 CPE3/CPE4 elements, to account for stress concentrations between closely spaced fibers and at fiber-matrix interfaces. The finely meshed microstructure is converted into a \(128 \times 128\) pixel image format by assigning a binary value (fiber or matrix) to each pixel, according to the dominant phase (by area fraction). The associated stress output is similarly converted to a \(128 \times 128\) pixel image by averaging the stresses within each pixel area.  

\begin{figure}[!ht]
    \centering
    \includegraphics[width=\textwidth]{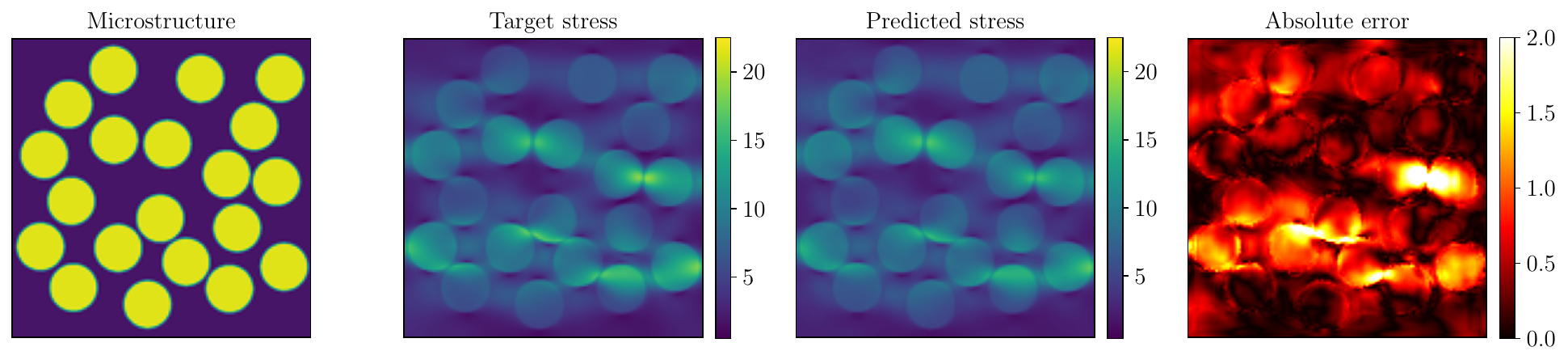}
    \caption{Stress field prediction using a standard deterministic U-net for one representative microstructure in the fiber-reinforced composite test dataset. Absolute error refers to the absolute value of the difference between the stresses predicted by FE and U-net.}
    \label{Deterministic_fiber_5}
\end{figure}

A standard deterministic U-net is first defined having parameters given in Table \ref{U_net_architecture_table} of the \hyperref[Appendix]{Appendix}. The model is trained using the entire dataset in a single batch, with an initial learning rate \(l_r=0.005\) for \(E = 1000\) epochs. The results for one representative image from the test dataset are presented in Fig.~\ref{Deterministic_fiber_5}. These results show that the U-net prediction yields reasonable accuracy with respect to the FEM solution, having an overall mean absolute prediction error \(\mu_{\boldsymbol{\varepsilon}} = 0.4562\). Similar results are found for the remaining images in the test dataset. 
 
\begin{figure}[!ht]
    \centering
    \includegraphics[width = 0.8\textwidth]{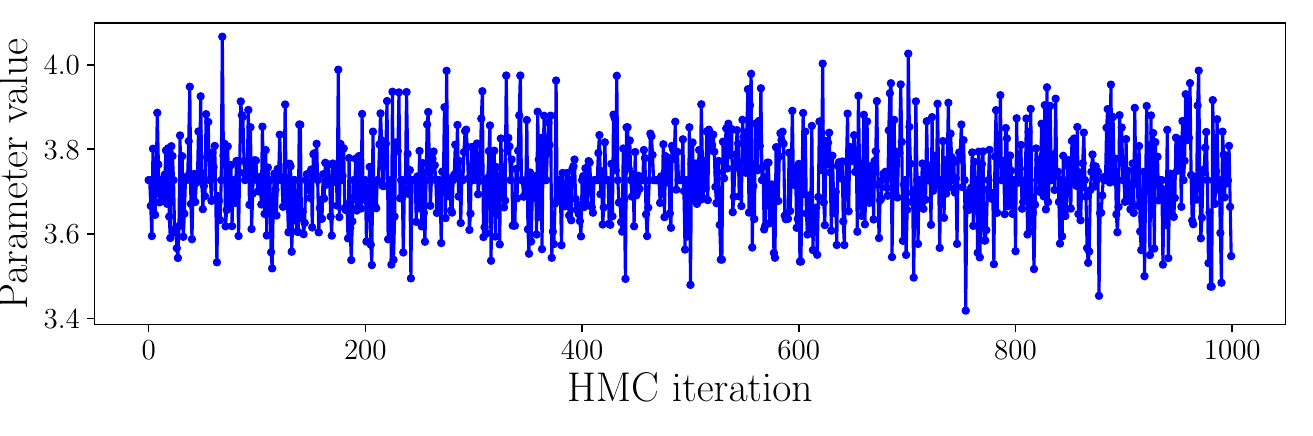}
    \includegraphics[width = 0.8\textwidth]{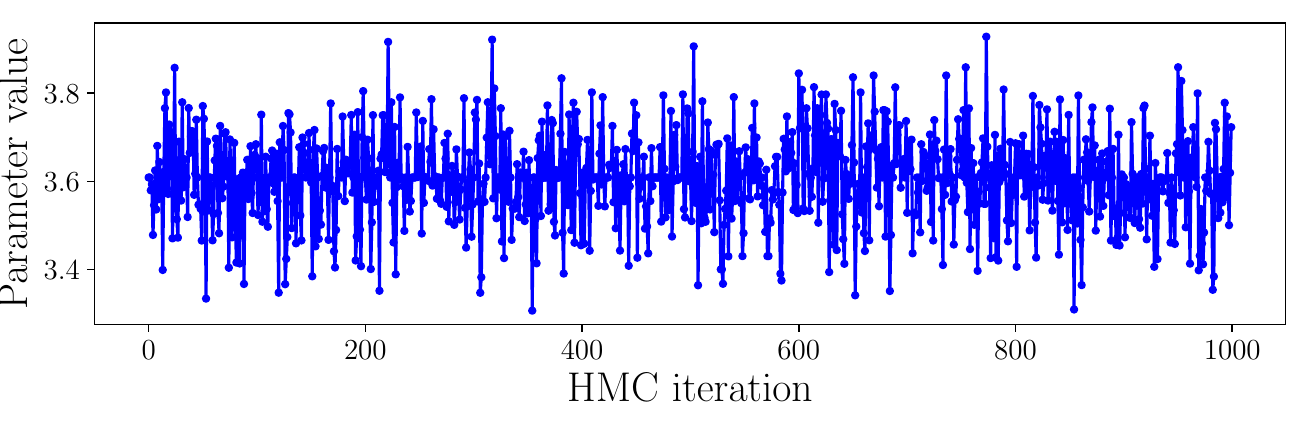}
    \caption{Trace plots for two representative NN parameters for a total of \(N_s=1000\) HMC samples.}
    \label{HMC_trajectories}
\end{figure}

Next, the pre-trained deterministic model was used to initiate the HMC algorithm for a fully Bayesian implementation. The network is sampled for a total of \(N_s=1000\) HMC samples, using step size \(\varepsilon = 0.0005\) and  trajectory length \(L = 300\). The noise variance is set to \(\sigma_{\alpha}^2 = 0.01\) and the prior variance is set to \(\sigma_{prior}^2 = 0.1\). As a simple diagnostic, Fig.~\ref{HMC_trajectories} shows trace plots for two representative NN parameters demonstrating that the autocorrelation between subsequent HMC samples is low, and indicating that the algorithm explores the state space effectively.

\begin{figure}[!ht]
    \begin{subfigure}{\textwidth}
    \centering
    \includegraphics[width=\linewidth]{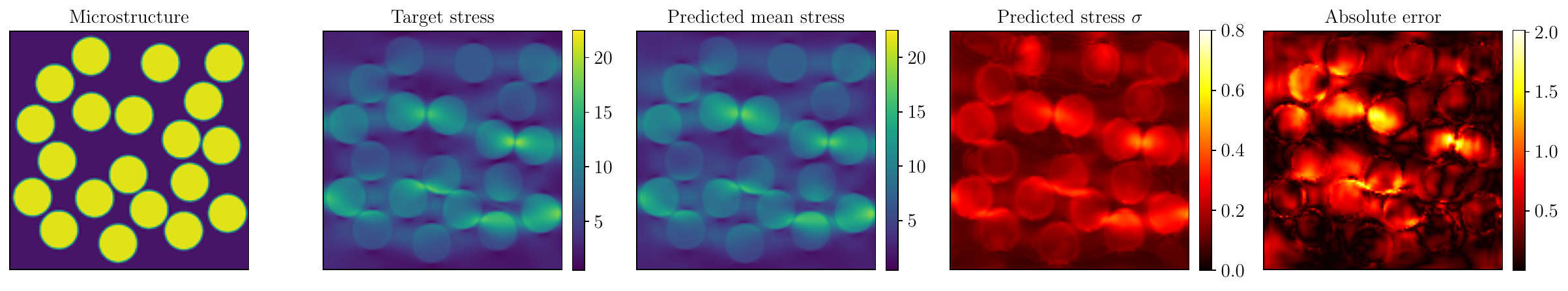}
    \caption{}
    \label{HMC_fiber_5}
    \end{subfigure}
    \begin{subfigure}{\textwidth}
    \centering
    \includegraphics[width=\textwidth]{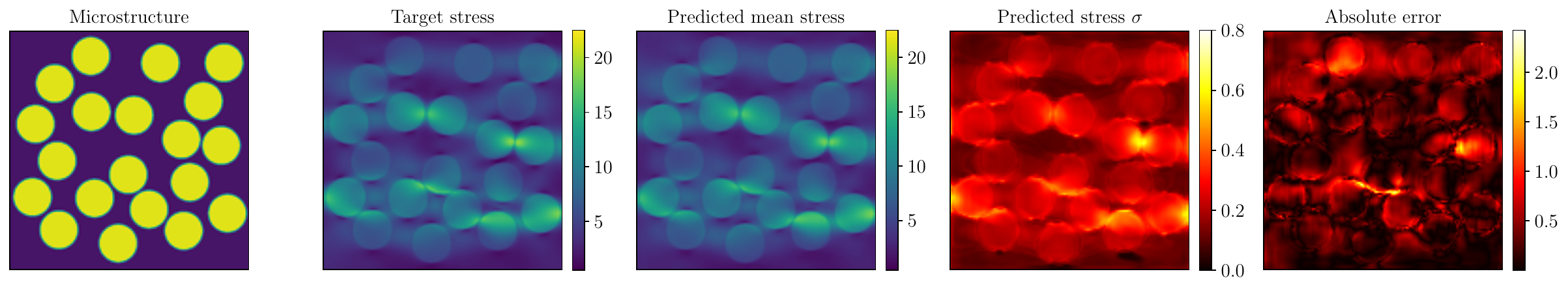}
    \caption{}
    \label{BBB_fiber_5}
    \end{subfigure}
    \begin{subfigure}{\textwidth}
    \centering
    \includegraphics[width=\textwidth]{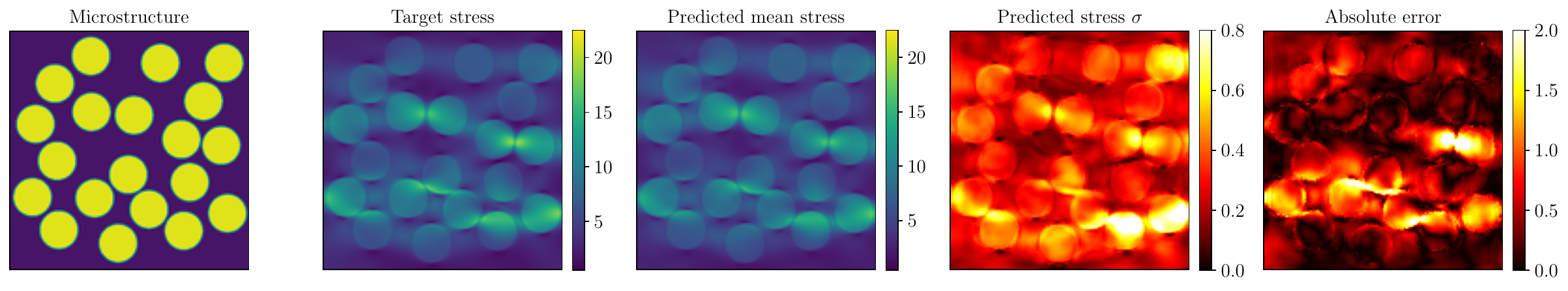}
    \caption{}
    \label{MCD_fiber_5}
    \end{subfigure}
    \caption{Mean stress field predictions, associated uncertainties, and absolute errors for the mean prediction from a Bayesian U-net for one representative microstructure in the fiber-reinforced composite test dataset: (a) HMC implementation, (b) BBB implementation, (c) MCD implementation.}
\end{figure}

 Fig.~\ref{HMC_fiber_5} shows the mean prediction, standard deviation, and absolute error relative to the FE-based values for one representative image in the test dataset using HMC. Comparing Fig.~\ref{Deterministic_fiber_5} and Fig.~\ref{HMC_fiber_5}, and specifically the absolute reconstruction error compared to the target FEM simulation, we see that the mean reconstruction from HMC exhibits a  higher degree of accuracy than the deterministic network prediction, while also providing a measure of predictive uncertainty at every pixel in the image. As anticipated, higher uncertainty is observed in regions of high stress and in areas near the fiber-matrix interfaces, where data discretization discrepancies are more likely. Due to the small degree of aleatoric uncertainty in the data, only the epistemic uncertainty is plotted.

To assess the BBB algorithm, the Bayesian U-net was also trained for \(E = 1000\) epochs using the following parameters: \(\lambda = 10^{-8}\), \(\sigma_{prior}^2 = 0.1\), and \(\mathbf{m}_{\mathbf{\omega}} = \mathbf{0}\). Note that, unlike HMC where the cost is very high and pre-training was deemed necessary, the BBB U-net was not initialized from the deterministic U-net.
Following the training, we sampled the distributions of the Bayesian convolutional layers to generate \(N_s = 1000\) realizations of the network. The results were then used to estimate the mean and the predictive variance using Eqs.~\eqref{BBB_mean} and \eqref{BBB_variance}. The mean stress prediction, associated variance, and absolute error for the same representative sample are shown in Fig.~\ref{BBB_fiber_5} for BBB. 
Similar to HMC, the BBB algorithm yields enhanced mean stress prediction accuracy compared to the deterministic network and, although the estimates of the standard deviation are higher than those predicted from HMC, the regions of high uncertainty appear to be consistent with HMC. 

Applying the MCD algorithm, the deterministic pre-trained U-net is augmented with dropout layers, which are sampled \(N_s=1000\) times to generate the predictive mean and variance using Eqs.~\eqref{MCD_mean} and \eqref{MCD_variance}, respectively. Given the heuristic nature of MCD, and considering that the posterior distribution may have arbitrarily small or large variance depending on the dropout rate and the model size \cite{osband2016risk}, a systematic study of the effects of different configurations of dropout layers and dropout rates was performed. Following a similar approach to Kendall et al.~\cite{kendall2015bayesian} the following cases were studied:
\begin{itemize}
    \item Case 1: Dropout in the innermost layer of the encoding path
    \item Case 2: Dropout in the innermost layer of the decoding path
    \item Case 3: Dropout layers in the innermost two layers of the encoding path
    \item Case 4: Dropout layers in the innermost two layers of the decoding path
\end{itemize}
where, for each case, the following dropout rates were employed: \(p = \{0.01, 0.05, 0.1, 0.2\}\). This amounts to sixteen distinct scenarios, and the the results with the lowest stress mean absolute error and stress error variance (Case 3 having $p=0.2$) are shown in Fig.~\ref{MCD_fiber_5} for the same representative sample. We see from Fig.~\ref{MCD_fiber_5} that MCD yields lower accuracy in the mean stress field prediction than HMC and BBB, and produces epistemic uncertainty predictions that are largely inconsistent with those from HMC and BBB. That is, they exhibit large uncertainties in different regions than HMC and BBB and high uncertainties do not necessarily spatially correspond to regions of larger mean stress or stress concentrations. 

\begin{figure}[!ht]
    \centering
    \includegraphics[width=\textwidth]{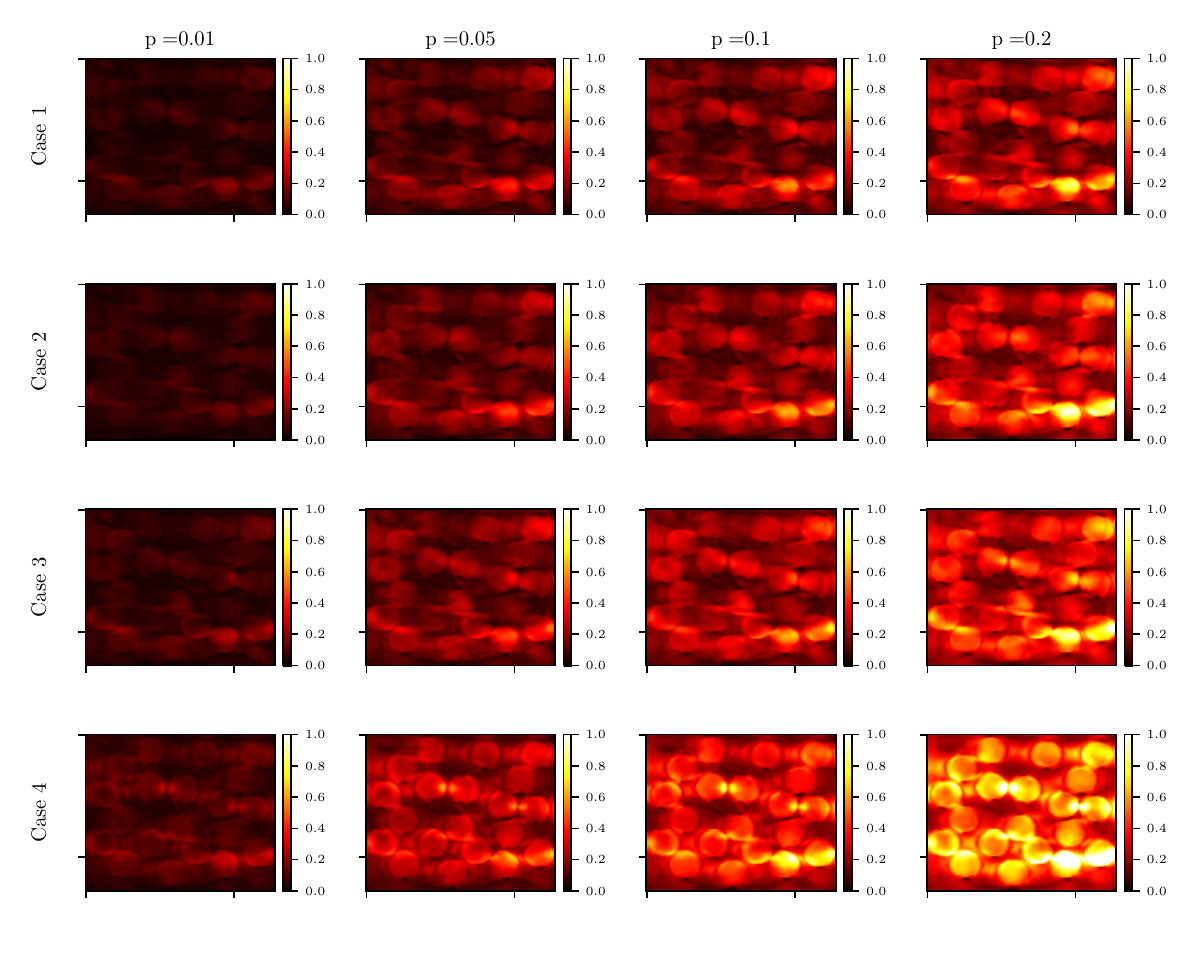}
    \caption{Estimated epistemic uncertainty fields using the MCD algorithm with different combinations of dropout layer locations and dropout rates for a representative microstructure in the fiber-reinforced composite test dataset.}
    \label{fiber_MCD_std_comparison}
\end{figure}

The resulting epistemic uncertainty fields for all 16 MCD scenarios are plotted in Fig.~\ref{fiber_MCD_std_comparison} for the same representative microstructure as in Fig.~\ref{MCD_fiber_5}. 
As expected, increasing the dropout rate leads to higher standard deviation in the prediction, but does not drastically change the spatial distribution of uncertainty. In contrast, the location of the dropout layers in the network has a more pronounced effect on the spatial distribution of uncertainty. This highlights the undesirable sensitivity of the MCD method to arbitrary parameter selections that yield inconsistent, and arguably unreliable uncertainty predictions. 

A comprehensive comparison of HMC, BBB, and the various MCD scenarios is provided in Table \ref{parametric_comparison_fiber} using the error metrics introduced in Section \ref{Data_and_metrics} across all samples in the test dataset. Specifically, the \(\mu_{\boldsymbol{\varepsilon}}\) in Eq.~\eqref{mae} and \(\sigma_{\boldsymbol{\varepsilon}}\) in Eq.~\eqref{error_variance} are obtained by comparing the predictions relative to the FE-based stress fields whereas  \(\sigma_{avg}\) in Eq.~\eqref{s_avg} is obtained by ensemble averaging. Further, \(\mu_{\boldsymbol{\varepsilon}_{\hat{\boldsymbol{\sigma}}}}\) in Eq.~\eqref{variance_mae} and \(\sigma_{\boldsymbol{\varepsilon}_{\hat{\boldsymbol{\sigma}}}}\) in Eq.~\eqref{variance_error_variance} are derived by comparing the variance predictions with the HMC-based results.

\begin{table}[!ht]
    \centering
    \begin{tabular}{lccccc}
        \toprule
          Case & \(\mu_{\boldsymbol{\varepsilon}}\) & \(\sigma_{\boldsymbol{\varepsilon}}\) & \(\sigma_{avg}\) & \(\mu_{\boldsymbol{\varepsilon}_{\hat{\boldsymbol{\sigma}}}}\) & \(\sigma_{\boldsymbol{\varepsilon}_{\hat{\boldsymbol{\sigma}}}}\) \\
        \midrule
        HMC & \bf{0.3856} & \bf{0.3948} & \bf{0.1693} & - & - \\
        BBB & \bf{0.4042} & \bf{0.4654} & \bf{0.2084} & \bf{0.0444} & \bf{0.0330} \\
        MCD-1, $p=0.01$ & 0.4667 & 0.5452 & \textbf{0.05820} & 0.1111 & 0.0578 \\
        MCD-1, $p=0.05$ & 0.4630 & 0.5390 & 0.1304 & 0.0539 & 0.0390 \\
        MCD-1, $p=0.1$ & 0.4591 & 0.5308 & 0.1829 & 0.0500 & 0.0481 \\
        MCD-1, $p=0.2$ & 0.4528 & 0.5169 & 0.2589 & 0.0960 & 0.0942 \\
        MCD-2, $p=0.01$ & 0.4674 & 0.5463 & 0.0673 & 0.1021 & 0.0530 \\
        MCD-2, $p=0.05$ & 0.4663 & 0.5439 & 0.1524 & 0.0500 & 0.0472 \\
        MCD-2, $p=0.1$ & 0.4654 & 0.5409 & 0.2192 & 0.0649 & 0.0933 \\
        MCD-2, $p=0.2$ & 0.4643 & 0.5367 & 0.3206 & 0.1529 & 0.1599 \\
        MCD-3, $p=0.01$ & 0.4663 & 0.5449 & 0.0769 & 0.0924 & 0.0539 \\
        MCD-3, $p=0.05$ & 0.4616 & 0.5375 & 0.1721 & \bf{0.0422} & \bf{0.0372} \\
        MCD-3, $p=0.1$ & 0.4572 & 0.5288 & 0.2441 & 0.0809 & 0.0723 \\
        MCD-3, $p=0.2$ & \bf{0.4517} & \bf{0.5124} & 0.34880 & 0.1803 & 0.1184 \\
        MCD-4, $p=0.01$ & 0.4674 & 0.5462 & 0.1064 & 0.0632 & 0.0398 \\
        MCD-4, $p=0.05$ & 0.4665 & 0.5452 & 0.2414 & 0.0755 & 0.0633 \\
        MCD-4, $p=0.1$ & 0.4663 & 0.5424 & 0.3477 & 0.1789 & 0.1159 \\
        MCD-4, $p=0.2$ & 0.4690 & 0.5426 & 0.5097 & 0.3405 & 0.1970 \\
        \bottomrule
    \end{tabular}    
    \caption{Error metric comparison across all samples in the test fiber reinforced composite dataset for the HMC, BBB and MCD methods: \(\mu_{\boldsymbol{\varepsilon}}=\) mean absolute prediction error (\(2^{nd}\) column - Eq.~\eqref{mae}) and \(\sigma_{\boldsymbol{\varepsilon}}=\) prediction error standard deviation (\(3^{rd}\) column - Eq.~\eqref{error_variance}) relative to the FE-based stress fields; \(\sigma_{avg}\) average standard deviation (\(4^{th}\) column - Eq.~\eqref{s_avg}); \(\mu_{\boldsymbol{\varepsilon}_{\hat{\boldsymbol{\sigma}}}} =\) mean HMC-relative variance error (\(5^{th}\) column - Eq.~\eqref{variance_mae}); \(\sigma_{\boldsymbol{\varepsilon}_{\hat{\boldsymbol{\sigma}}}} =\) HMC-relative variance error standard deviation (\(6^{th}\) column - Eq.~\eqref{variance_error_variance}). The best estimates for each method are given in boldface text.}
    \label{parametric_comparison_fiber}
\end{table}

Clearly, the HMC method yields the lowest mean absolute prediction error (\(\mu_{\boldsymbol{\varepsilon}}\)) and prediction error standard deviation (\(\sigma_{\boldsymbol{\varepsilon}}\)). BBB meanwhile, exhibits slightly higher values of mean absolute prediction error and prediction error standard deviation than HMC but it yields considerably more accurate results when compared to all of the MCD cases. The BBB and the MCD methods generate comparably accurate results with respect to mean HMC-relative variance error (\(\mu_{\boldsymbol{\varepsilon}_{\hat{\boldsymbol{\sigma}}}}\)) and HMC-relative variance error standard deviation (\(\sigma_{\boldsymbol{\varepsilon}_{\hat{\boldsymbol{\sigma}}}}\)) when compared with the HMC-generated results. Of course, the optimal results from MCD are only found by picking the optimal case of 16 different sets of hyperparameters, while the results from BBB are based on only one analysis. 

\begin{figure}[!ht]
    \centering
    \includegraphics[width=0.7\textwidth]{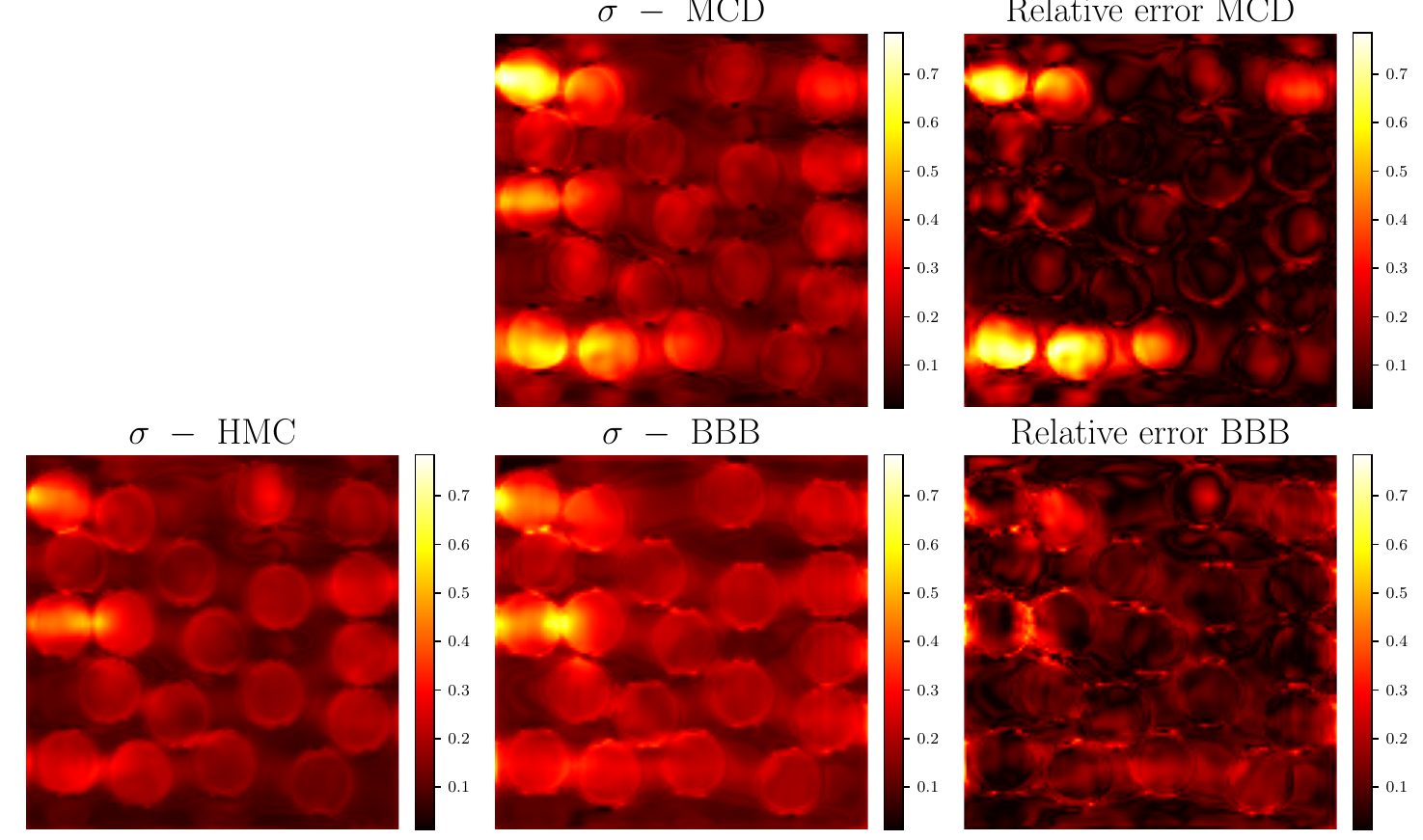}
    \caption{Epistemic uncertainty estimates for one representative microstructure in the fiber-reinforced composite test dataset using the HMC, MCD and BBB algorithms. The relative error is calculated using HMC as the target variance estimate.} 
    \label{std_comparison_4}
\end{figure}

The results suggest that HMC is the most accurate but computationally expensive method for performing UQ in stress field prediction for the fiber reinforced composite material, which is consistent with expectations.
Using HMC as the benchmark, Fig.~\ref{std_comparison_4} shows the corresponding epistemic uncertainty fields for all three methods and the error fields between the HMC and the BBB, MCD-based results. The MCD configuration was chosen as Case 3 with $p=0.2$, which gives the the lowest mean absolute prediction error (\(\mu_{\boldsymbol{\varepsilon}}\)) and prediction error standard deviation (\(\sigma_{\boldsymbol{\varepsilon}}\)) based on the results given in Table~\ref{parametric_comparison_fiber}.
Further, there is no unique MCD configuration that consistently generates accurate results across all the error metrics. In fact, the BBB algorithm -- which requires minimal tuning compared to HMC and MCD, as well as less computational time than the HMC method -- outperforms almost all cases of MCD. 

As a note on computational performance, training and predictions were performed using the PyTorch ML library with CUDA acceleration in which the training process has been carried out on one NVIDIA A100 GPU with 80GB of memory using 16 tasks per node. The MCD and BBB
approaches require approximately 2h of computation time for training and can also be performed on a standard desktop/laptop computer. In comparison, the HMC methodology requires approximately 66.5 h of computation time for training. Considering the above, the BBB algorithm is significantly more robust than MCD method and yields a reasonably accurate representation of uncertainty while offering a significant computational speedup over HMC.

\subsection{Polycrystalline material}
In the next numerical example, the behavior of a two-dimensional polycrystalline material under tensile load is studied. The initial microstructure geometries are synthesized with the aid of the MicroStructPy package \cite{hart2020microstructpy}, using the following steps: an accelerated seed placement with overlapping grains, a spherical grain decomposition procedure, Voronoi tessellation and mesh generation. A total of \(n=1140\) diverse microstructures are generated using the following parameters: \(number \ of \ grains = 20, \ side \ length = 10,\ grain\ size = 1, \ grain\ variance = 1\) where a high variance-to-grain size ratio is chosen to obtain statistically representative samples. The following orthotropic elastic properties are assigned to each grain: \(C_{1111} = C_{2222} = C_{3333} = 204.6 \text{ GPa}\), \(C_{1122} = C_{1133} = C_{2233} = 137.7 \text{ GPa}\), \(C_{1212} = C_{1313} = C_{2323} = 126.2 \text{ GPa}\) where \(C_{ijkl}\) denote the parameters of the elasticity tensor in the local crystallographic direction of the grain. Following the generation of the mesh, a unique random orientation with values between \(-\pi\) and \(\pi\) is assigned to each grain, and the elastic properties are transformed according to that grain's crystallographic orientation. The elastic stresses associated with the microstructure are generated using the ABAQUS \cite{standard1995user} software, in which a unit displacement boundary condition is applied to the right hand side (with fixed displacement on the left hand side) and traction-free boundary conditions are applied to the top and bottom boundaries. The mesh for the finite element analyses contains $\approx$19400 CPE3 elements per microstructure sample.  The first stress component, \(\sigma_{11}\), is taken as the output using a single channel image. The final input-output pairs are discretized into \(128 \times 128\) pixels, following the same procedure used for the fiber-reinforced composite example. 

\begin{figure}[!ht]
    \centering
    \includegraphics[width=\textwidth]{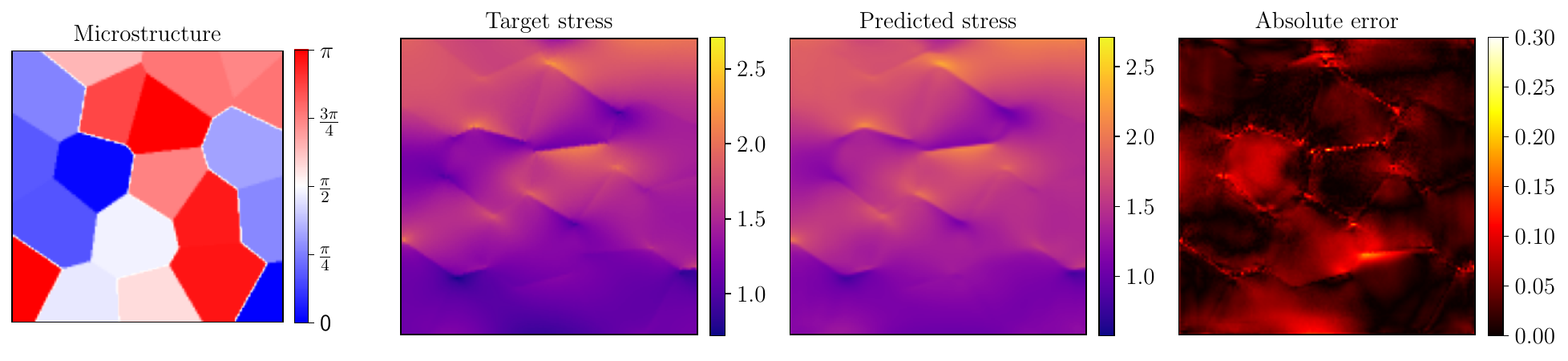}
    \caption{Stress field prediction using the deterministic U-net architecture for one representative microstructure in the polycrystalline material test dataset.}
    \label{Deterministic_polycrystalline_9}
\end{figure}

The reader is referred to Table \ref{U_net_architecture_table} of the \hyperref[Appendix]{Appendix} for a detailed summary of the model architecture. The model is trained using the same parameters as the previous application. Specifically, a single batch of training data is used with an initial learning rate \(l_r=0.005\) for a total of \(E = 1000\) epochs. The stress field prediction for a representative sample in the test dataset using a deterministic U-net is plotted in Fig.~\ref{Deterministic_polycrystalline_9} along with the stress computed using FEM and the absolute error field.
The FEM stress field is accurately estimated by the U-net with mean absolute error \(\mu_{\boldsymbol{\varepsilon}} = 0.03858\) across all the samples in the test dataset. This is encouraging considering that the input material microstructure representation contains real values between \(-\pi\) and \(\pi\), which renders the problem more challenging compared to the binary valued fiber-reinforced composite case. Further, although the stress is generated solving a linear elastic simulation, the relationship between grain orientation and stress amplitude is highly complex.

\begin{figure}[!ht]
    \begin{subfigure}{\textwidth}
        \centering
        \includegraphics[width=\linewidth]{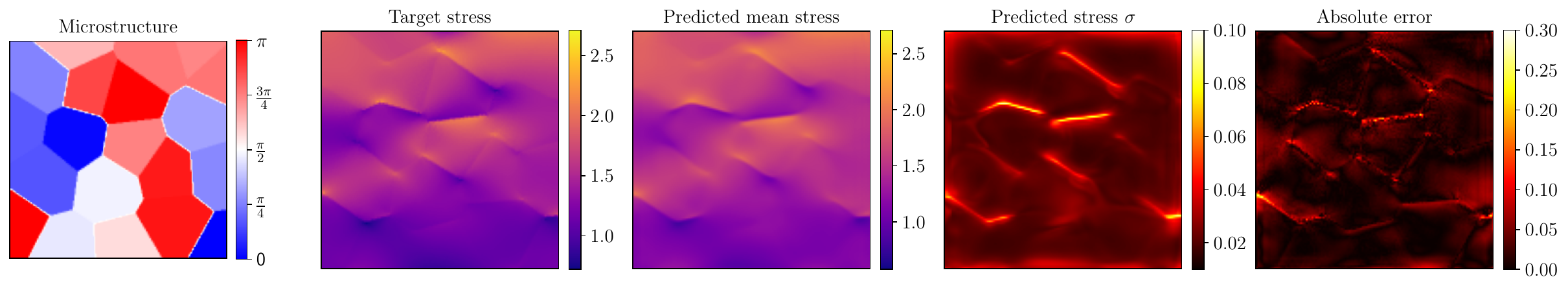}
        \caption{}
        \label{BBB_polycrystalline_9}
    \end{subfigure}
    \begin{subfigure}{\textwidth}
        \centering
        \includegraphics[width=\linewidth]{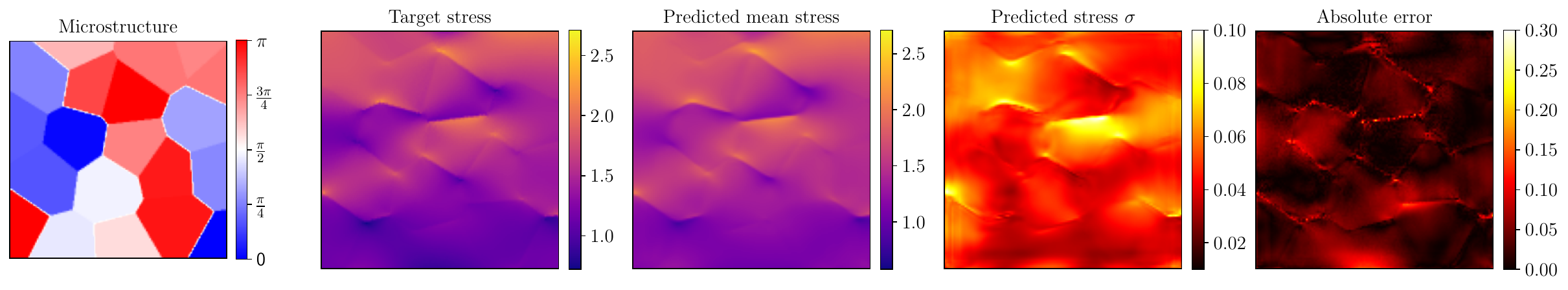}
        \caption{}
        \label{MCD_polycrystalline_9}
    \end{subfigure}
        \caption{Stress field prediction using BBB-generated samples of the U-net architecture for one representative microstructure in the polycrystalline material test dataset.}
\end{figure}

The aforementioned complexities render the application of HMC infeasible for several reasons. First, the size of the convolutional kernel (\(k=9\)) increases the complexity of the numerical gradient evaluation due to the high number of parameters without model pre-training. Second, the highly complex input-output relationship requires a considerably higher trajectory length \(L>300\) which significantly increases the computational time. For this reason, and considering its performance for the composite material, the BBB algorithm is applied as a benchmark to quantify prediction uncertainty. The parameters for BBB are the same as the previous example: \(E = 1000\) epochs, \(\lambda = 10^{-8}\), prior variance \(\sigma_{prior}^2 = 0.1\), prior mean \(\mathbf{m}_{\mathbf{\omega}} = \mathbf{0}\), and \(M=1\) MCS samples of the variational posterior per iteration. The distributions of the trained convolutional parameters are sampled \(N_s = 1000\) times, where the predictive mean and predictive epistemic uncertainty are calculated using Eqs.~\eqref{BBB_mean} and \eqref{BBB_variance}. The results for a representative sample in the test dataset are plotted in Fig.~\ref{BBB_polycrystalline_9}. 
Comparing the absolute error between Fig.~\ref{Deterministic_polycrystalline_9} and Fig.~\ref{BBB_polycrystalline_9}, the BBB algorithm modestly enhances accuracy relative to the deterministic network while quantifying epistemic uncertainty. As expected, areas of higher uncertainty are concentrated primarily in the grain boundaries where the prediction accuracy decreases.

\begin{figure}[!ht]
    \centering
    \includegraphics[width=\textwidth]{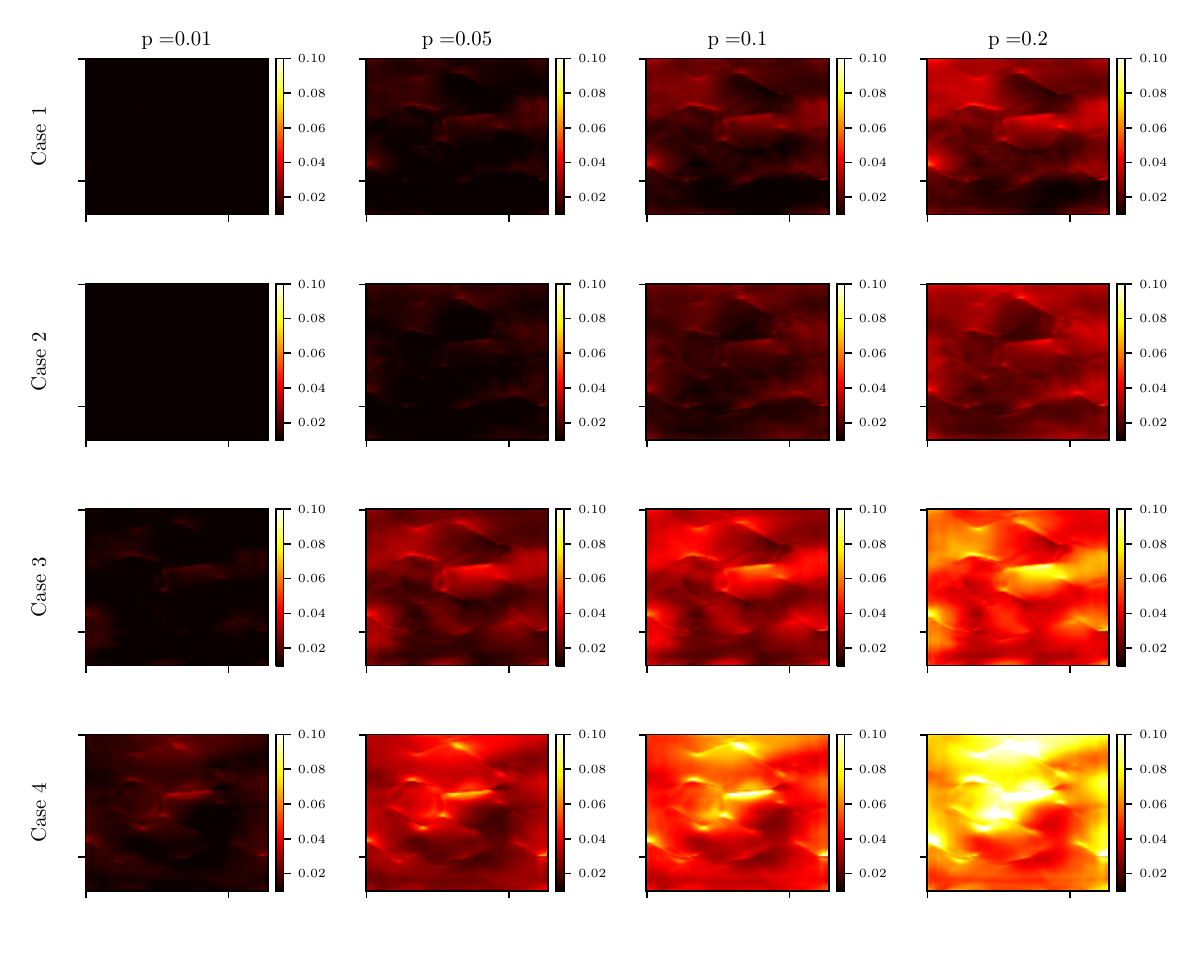}
    \caption{Estimated epistemic uncertainty fields using the MCD algorithm with different combinations of dropout layer locations and dropout rates for a representative microstructure in the polycrystalline material test dataset.}
    \label{polycrystalline_MCD_std_comparison}
\end{figure}

Next, the performance of the MCD-dropout algorithm is assessed. Dropout layers are introduced in the pre-trained U-net, for which the underlying Bernoulli distribution is sampled \(N_s=1000\) times to evaluate the expressions Eqs.~\eqref{MCD_mean} and \eqref{MCD_variance}. MCD is again investigated for each of the cases presented in the previous numerical example. The results of the configuration (Case 3, \(p=0.2\)) that yields the lowest mean absolute prediction error and prediction
error standard deviation are presented in Fig.~\ref{MCD_polycrystalline_9} for the same representative microstructure. MCD produces mean predictions of comparative accuracy when compared to BBB, but with errors concentrated in different regions. For example, MCD exhibits higher error within the grains. Moreover, the quantified uncertainties differ significantly. Comparison of the epistemic uncertainty fields for all 16 MCD configurations are shown in Fig.~\ref{polycrystalline_MCD_std_comparison}.
The results are consistent with the fiber-reinforced composite case. In particular, it can be argued that the dropout rate largely determines the magnitude of uncertainty, while the positions and the number of dropout layers influences its distribution.

To summarize, Table~\ref{parametric_comparison_polycrystalline} gives the evaluated error metrics introduced in Section \ref{Data_and_metrics} across all the samples in the test dataset for BBB and all of the MCD configurations. Comparison to HMC is not provided, again due to its infeasibility. Specifically, \(\mu_{\boldsymbol{\varepsilon}}\) in Eq.~\eqref{mae} and \(\sigma_{\boldsymbol{\varepsilon}}\) in Eq.~\eqref{error_variance} are calculated by comparing the predictions relative to the FE-based stress fields whereas  \(\sigma_{avg}\) in Eq.~\eqref{s_avg} is obtained as the ensemble average of the variances. 

\begin{table}[!ht]
    \centering
    \begin{tabular}{lccc}
        \toprule
        Case & \(\mu_{\boldsymbol{\varepsilon}}\) & \(\sigma_{\boldsymbol{\varepsilon}}\) & \(\sigma_{avg}\) \\
        \midrule
        BBB & \textbf{0.0241} & \textbf{0.0227} & \textbf{0.02310} \\
        MCD-1, $p=0.01$ & 0.0385 & 0.0317 & \textbf{0.0052} \\
        MCD-1, $p=0.05$ & 0.0381 & 0.0314 & 0.0117 \\
        MCD-1, $p=0.1$ & 0.0377 & 0.0311 & 0.0167 \\
        MCD-1, $p=0.2$ & 0.0371 & 0.0307 & 0.0243 \\
        MCD-2, $p=0.01$ & 0.0385 & 0.0317 & 0.0055 \\
        MCD-2, $p=0.05$ & 0.0384 & 0.0316 & 0.0124 \\
        MCD-2, $p=0.1$ & 0.0383 & 0.0315 & 0.0179 \\
        MCD-2, $p=0.2$ & 0.0381 & 0.0313 & 0.0264 \\
        MCD-3, $p=0.01$ & 0.0384 & 0.0316 & 0.0103 \\
        MCD-3, $p=0.05$ & 0.0376 & 0.031 & 0.0231 \\
        MCD-3, $p=0.1$ & 0.0371 & 0.0304 & 0.0328 \\
        MCD-3, $p=0.2$ & \textbf{0.0362} & \textbf{0.0296} & 0.04770 \\
        MCD-4, $p=0.01$ & 0.0385 & 0.0316 & 0.01380 \\
        MCD-4, $p=0.05$ & 0.0383 & 0.0313 & 0.03110 \\
        MCD-4, $p=0.1$ & 0.0384 & 0.0313 & 0.04450 \\
        MCD-4, $p=0.2$ & 0.0394 & 0.0322 & 0.06470 \\
        \bottomrule
    \end{tabular}
    \caption{Error metric comparison across all samples in the test polycrystalline microstructure dataset for the BBB and MCD methods: \(\mu_{\boldsymbol{\varepsilon}}=\) mean absolute prediction error (\(2^{nd}\) column - Eq.~\eqref{mae}) and \(\sigma_{\boldsymbol{\varepsilon}}=\) prediction error standard deviation (\(3^{rd}\) column - Eq.~\eqref{error_variance}) relative to the FE-based stress fields; \(\sigma_{avg}\) average standard deviation (\(4^{th}\) column - Eq.~\eqref{s_avg});.  The best estimates for each method are given in boldface text.
    \label{parametric_comparison_polycrystalline}}
\end{table}

The BBB outperforms all cases of MCD by producing more accurate results (lower \(\mu_{\boldsymbol{\varepsilon}}\)) with a smaller prediction error standard deviation \(\sigma_{\boldsymbol{\varepsilon}}\)  and average standard deviation \(\sigma_{avg}\). Interestingly, similar to the composite material example, the MCD configuration that yields the lowest prediction error is case 3 (dropout at the innermost two layers of the encoding path) with drop rate \(p=0.2\) despite the fact that the network architecture differs significantly (e.g. kernel size and number of filters) and the input-output relationship is more complex.  

\section{Concluding remarks}
In this paper, the predictive capabilities of Bayesian convolutional neural networks for two-dimensional stress field estimation with uncertainty are explored. This assessment is performed by considering a fiber-reinforced composite and a polycrystalline material system in conjunction with three representative deep learning-based UQ methodologies which exhibit distinct trade-offs between computational complexity and accuracy. These are the posterior sampling-based HMC method, and the variational MCD and BBB algorithms. Overall, it is shown that a probabilistic treatment of the mapping between the microstructure and the stress yields enhanced (or comparable) predictive accuracy in the mean value predictions, as compared to the standard deterministic implementation. The Bayesian methods also provide estimates of the epistemic uncertainty in predictions, helping to quantify the degree of confidence in the predictions across the entire spatial domain, distinguishing regions of high prediction uncertainty (e.g. near grain boundaries or fiber matrix interfaces) and those of relatively low prediction uncertainty. Further, the applicability of HMC as a reference UQ methodology is demonstrated for the first time in such a high-dimensional materials setting, where it is systematically compared to the other, less computationally intensive UQ methods. As anticipated, it is found that the MCD technique is sensitive to the model size, the dropout layer and dropout rate configuration and may yield inconsistent UQ results. In contrast, the BBB algorithm strikes a good balance between computational time, predictive accuracy and UQ robustness. It is also noted that the proposed framework can also be applied as a general UQ methodology in diverse scientific applications where the task relates to two-dimensional full-field estimation. It is not limited to stress field prediction.

The current study has primarily focused on quantifying the epistemic uncertainty within a surrogate modeling framework, where the aleatoric uncertainty is assumed to be negligible and homoscedastic as a result of the synthetically generated microstructures. Therefore, generalizing the approach to account for more complex experimental data endowed with considerable levels of aleatoric noise is a natural extension. Finally, to improve the feasibility of HMC and yield better uncertainty estimates, the use of parallel Markov chains \cite{calderhead2014general} and more advanced algorithms such as the No-U-Turn Sampler (NUTS) \cite{hoffman2014no} or Hamiltonian Neural Networks \cite{Dhulipala_Che_Shields_2023a} are potential research directions to better explore the probability space in a more coherent and computationally efficient manner.

\section{Declaration of Competing Interest}
The authors declare no competing interests.

\section{Acknowledgements}
Research was sponsored by the Army Research Laboratory and was accomplished under Cooperative Agreement Number W911NF-22-2-0121. The views and conclusions contained in this document are those of the authors and should not be interpreted as representing the official policies, either expressed or implied, of the Army Research Laboratory or the U.S. Government. The U.S. Government is authorized to reproduce and distribute reprints for Government purposes notwithstanding any copyright notation herein. This work was carried out at the Advanced Research Computing at Hopkins (ARCH) core facility  (rockfish.jhu.edu), which is supported by the National Science Foundation (NSF) grant number OAC1920103. The authors gratefully acknowledge Dr. Ashwini Gupta for providing the data and valuable insights regarding the fiber-reinforced composite material. 

\section{Appendix}\label{Appendix}
\begin{table}[htbp]
    \centering
    \begin{tabular}{@{}C{2cm}C{3cm}C{3cm}C{2.5cm}C{2cm}@{}}
    \toprule
    \bf{Operations} & \bf{Number of Filters} & \bf{Kernel Size} & \bf{Output Size} 
    \\
    Input microstructure & - & - & 1x128x128 \\
    Encoder block 1 & 16 & \begin{tabular}{@{}c@{}}3x3 (Fiber)\\9x9 (Polycrystalline)\end{tabular} & 16x128x128 \\
    Encoder block 2 & 64 & \begin{tabular}{@{}c@{}}3x3 (Fiber)\\9x9 (Polycrystalline)\end{tabular} & 64x64x64 \\
    Encoder block 3 & 128 & \begin{tabular}{@{}c@{}}3x3 (Fiber)\\9x9 (Polycrystalline)\end{tabular} & 128x32x32 \\
    Encoder block 4 & 256 & \begin{tabular}{@{}c@{}}3x3 (Fiber)\\9x9 (Polycrystalline)\end{tabular} & 256x16x16 \\
    Encoder block 5 & 512 & \begin{tabular}{@{}c@{}}3x3 (Fiber)\\9x9 (Polycrystalline)\end{tabular} & 512x8x8 \\
    Decoder block 1 & 256 & \begin{tabular}{@{}c@{}}3x3 (Fiber)\\9x9 (Polycrystalline)\end{tabular} & 256x16x16 \\
    Decoder block 2 & 128 & \begin{tabular}{@{}c@{}}3x3 (Fiber)\\9x9 (Polycrystalline)\end{tabular} & 128x32x32 \\
    Decoder block 3 & 64 & \begin{tabular}{@{}c@{}}3x3 (Fiber)\\9x9 (Polycrystalline)\end{tabular} & 64x64x64 \\
    Decoder block 4 & 32 & \begin{tabular}{@{}c@{}}3x3 (Fiber)\\9x9 (Polycrystalline)\end{tabular} & 32x128x128 \\
    Output layer & 1 & \begin{tabular}{@{}c@{}}3x3 (Fiber)\\9x9 (Polycrystalline)\end{tabular} & 1x128x128  \\
    \bottomrule
\end{tabular}
    \caption{U-net Network Architecture}
    \label{U_net_architecture_table}
\end{table}

\bibliographystyle{elsarticle-num}
\biboptions{sort&compress}
\bibliography{BCNN_materials.bib}
\end{document}